\documentclass[nohyperref]{article}

\usepackage{PRIMEarxiv}

\usepackage{hyperref}       % hyperlinks
\usepackage[utf8]{inputenc} % allow utf-8 input
\usepackage[T1]{fontenc}    % use 8-bit T1 fonts
\usepackage{url}            % simple URL typesetting
\usepackage{booktabs}       % professional-quality tables
\usepackage{amsfonts}       % blackboard math symbols
\usepackage{nicefrac}       % compact symbols for 1/2, etc.
\usepackage{microtype}      % microtypography
\usepackage{lipsum}
\usepackage{fancyhdr}       % header
\usepackage{graphicx}       % graphics
\graphicspath{{media/}}     % organize your images and other figures under media/ folder
\usepackage{subfigure}
\usepackage{booktabs} % for professional tables
\usepackage{amsfonts}
\usepackage{stfloats}
\usepackage{ulem}

\usepackage{amsmath}
\usepackage{adjustbox}
\usepackage{multirow}
\usepackage{array}
\usepackage{float}
\usepackage{caption}
\usepackage{amsmath}
\usepackage{amssymb}
\usepackage{mathtools}
\usepackage{amsthm}
\usepackage{balance}
\usepackage[capitalize,noabbrev]{cleveref}

\newcommand{\ofa}{OFA}

\makeatletter%用于连接公式编号 

\newenvironment{figurehere}
{\def\@captype{figure}}
{}
\makeatother%用于连接公式编号
  
\title{OFA: Unifying Architectures, Tasks, and Modalities Through a Simple Sequence-to-Sequence Learning Framework
}

\author{
  Peng Wang, An Yang, Rui Men, Junyang Lin, Shuai Bai\\\\
  \textbf{Zhikang Li, Jianxin Ma, Chang Zhou, Jingren Zhou, Hongxia Yang}\\\\
  DAMO Academy, Alibaba Group
  \thanks{Correspondence to: Chang Zhou<ericzhou.zc@alibaba-inc.com>.}
   \\\\
  \texttt{\{zheluo.wp, ya235025, menrui.mr, junyang.ljy, baishuai.bs,} \\
  \texttt{zhikang.lzk, jason.mjx, ericzhou.zc, jingren.zhou, yang.yhx\}@alibaba-inc.com} \\
}
\begin{document}
\maketitle

{
\begin{figurehere}
\centering
\includegraphics[width=1.\linewidth]{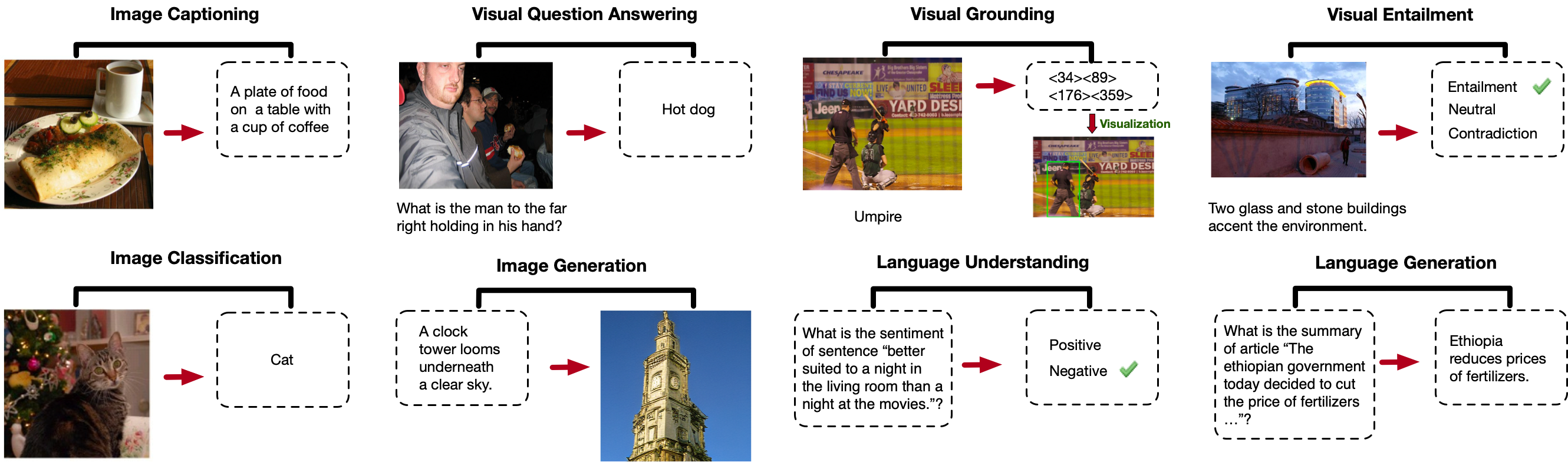}
\caption{Examples of various tasks supported by \ofa.}
\label{fig:examples}
% \vskip -0.2in
\end{figurehere}
}
\vskip .3in

\begin{abstract}
In this work, we pursue a unified paradigm for multimodal pretraining to break the scaffolds of complex task/modality-specific customization. 
% We propose \ofa, a unified multimodal pretrained model that unifies modalities (i.e., cross-modality, vision, language) and tasks (e.g., image generation, visual grounding, image captioning, image classification, text generation, etc.) to a simple sequence-to-sequence learning framework based on the encoder-decoder architecture. 
% OFA performs pretraining and finetuning with task instructions and introduces no extra task-specific layers for finetuning. 
We propose \ofa, a Task-Agnostic and Modality-Agnostic framework that supports Task Comprehensiveness. 
OFA unifies a diverse set of cross-modal and unimodal tasks, including image generation, visual grounding, image captioning, image classification, language modeling, etc., in a simple sequence-to-sequence learning framework.
%OFA unifies modalities (i.e., cross-modality, vision, language) and tasks (e.g., image generation, visual grounding, image captioning, image classification, text generation, etc.) to a simple sequence-to-sequence learning framework based on the encoder-decoder architecture. 
OFA follows the instruction-based learning in both pretraining and finetuning stages,  requiring no extra task-specific layers for downstream tasks. 
%It performs pretraining and finetuning with task instructions and introduces no extra task-specific layers for finetuning. 
% OFA is the first attempt to unify the vision \& language, vision-only and language-only tasks. 
In comparison with the recent state-of-the-art vision \& language models that rely on extremely large cross-modal datasets, OFA is pretrained on only $20$M publicly available image-text pairs. 
Despite its simplicity and relatively small-scale training data, OFA achieves new SOTAs in a series of cross-modal tasks
%including image captioning, visual question answering, visual entailment, referring expression comprehension, etc., 
while attaining highly competitive performances on uni-modal tasks. 
%, including language understanding, language generation, and image classification, 
% with state-of-the-art pretrained models.
%Experimental results show that OFA, pretrained on the publicly available datasets of $20$M image-text pairs, achieves state-of-the-art results in a series of multimodal tasks, including image captioning, visual question answering, visual entailment, referring expression comprehension, etc. 
%It also demonstrates comparable performance on single-modal tasks with state-of-the-art pretrained models, including language understanding, language generation, and image classification. 
%Our further analysis indicates that besides reaching superior performance in the finetuning scenario, OFA can effectively transfer to unseen tasks and unseen domains.
Our further analysis indicates that OFA can also effectively transfer to unseen tasks and unseen domains.
% Experimental results show that OFA achieves new state-of-the-arts on a series of multimodal tasks, including image captioning (COCO test CIDEr: 149.6), text-to-image generation (COCO test FID: 10.5), VQA (test-std acc.: 80.02), SNLI-VE (test acc.: 90.20), and referring expression comprehension (RefCOCO / RefCOCO+ / RefCOCOg test acc.: 92.93 / 90.10 / 85.20). 
% Through extensive analyses, we demonstrate that \ofa~reaches comparable performance with uni-modal pretrained models (e.g., BERT, MAE, MoCo v3, SimCLR v2, etc.) in uni-modal tasks, including NLU, NLG, and image classification, and it effectively transfers to unseen tasks and domains. 
Our code and models are publicly available at \url{https://github.com/OFA-Sys/OFA}.
\end{abstract}

% keywords can be removed
\keywords{Unified frameworks \and Multimodal pretraining \and Multitask learning \and Zero-shot learning}

\section{Introduction}
\label{sec:intro}
%The tremendous success of natural language pretraining has intrigued the fast development of multimodal pretraining. It had been challenging to learn high-quality cross-modal representation. 
%Multimodal pretraining, which integrates Transformer architecture and language modeling for vision \& language (VL) joint training, started a new era for this field. 
%Starting from 2019, a family of VL-transformer models~\cite{visualbert, vilbert, uniter, oscar, vinvl, m6, simvlm, vlmo} have created a series of new state-of-the-arts (SOTA) performance in the VL downstream tasks, including visual question answering (VQA), image captioning, etc. Another line of works take advantage of weakly supervised image-text data for retrieval~\cite{clip, align} and image synthesis~\cite{dalle, cogview, nvwa}. 
%The advances imply that large-scale multimodal pretraining is a promising direction towards a unified foundation model~\cite{foundation_model}.

Building an omnipotent model that handles as many tasks and modalities as human beings is an attractive goal in the AI community. 
%The central problem towards this goal is to represent massive varieties of modalities, tasks, and training regimes in a single model.
The possibilities of achieving this goal may largely depend on whether massive varieties of modalities, tasks and training regimes can be represented with only a few forms that can be unified and managed by a single model or system.

Recent developments of the Transformer~\cite{transformer} architecture have shown its potential for being a universal computation engine~\cite{bert,gpt3,transformer_math,wav2vec,vit,perceiver,vlbert}. 
% based on which the pretrain-finetune paradigm achieves great successes in many domains. 
% Language models, prompt~\cite{gpt3} and instruction~\cite{flan,t0} tuning regimes also demonstrate themselves to be powerful zero-shot learners.
In the settings of supervised learning, the ``pretrain-finetune'' paradigm achieves excellent success in many domains. In the regimes of few-/zero-shot learning, language models with prompt / instruction tuning prove powerful zero-/few-shot learners~\cite{gpt3, flan, t0}. 
These advances have provided more significant than ever opportunities for the emergence of an omni-model. 

To support better generalization for open-ended problems while maintaining multitask performance and ease of use, we advocate that an omnipotent model should have the following three properties:
%A question naturally arises: \textit{What should we do to make multimodal pretraining simple and generalizable?}
%We advocate building an omni-model to meet the following properties:
1. Task-Agnostic~(TA): unified task representation to support different types of tasks, including classification, generation, self-supervised pretext tasks, etc., and to be agnostic to either pretraining or finetuning.
2. Modality-Agnostic~(MA): unified input and output representation shared among all tasks to handle different modalities.
3. Task Comprehensiveness~(TC): enough task variety to accumulate generalization ability robustly. 

% However, satisfying the above three properties is challenging while achieving superior performance. 
However, it is challenging to satisfy these properties while maintaining superior performance in downstream tasks. 
Current language and multimodal pretrained models readily fail at parts of these properties, due to their following design choices.
1. Extra learnable components for finetuning, e.g., task-specific heads~\cite{bert}, adapters~\cite{adapter}, soft prompts~\cite{prompt_tuning}. 
This makes the model structure task-specific and poses discrepancy between pretraining and finetuning. Such designs are also not friendly to supporting unseen tasks in a zero-shot manner.
%It makes the model task-specific and thus violates TA. Besides, there are difficulties in finding an optimal component. Such designs are not friendly to supporting unseen tasks in a zero-shot manner.
2. Task-specific formulation. For most current methods, pretraining, finetuning and zero-shot tasks usually differ in task formulation and training objectives. This violates TA and it is burdensome to scale up the task population to achieve TC.
3. Entangling modality representation with downstream tasks. It is a common practice for Vision-Language models to take the detected objects as part of the image input features~\cite{vlbert, vilbert, uniter, oscar, villa, vinvl}. Though it demonstrates better downstream task performance on some closed-domain datasets, it depends on an extra object detector which usually fails at open-domain data.
Therefore, we explore an omni-model for multimodal pretraining and propose \textbf{\ofa}, hopefully ``One For All'', which achieves the objectives of unifying architectures, tasks, and modalities, and supports the three properties above.\footnote{This work is the latest one of our M6 series~\cite{m6, ufc, m6-t, m6-10t}.}
We formulate both pretraining and finetuning tasks in a unified sequence-to-sequence abstraction via handcrafted instructions~\cite{flan,t0} to achieve Task-Agnostic. 
A Transformer is adopted as the Modality-Agnostic compute engine, with a constraint that no learnable task- or modality-specific components will be added to downstream tasks. 
It is available to represent information from different modalities within a globally shared multimodal vocabulary across all tasks. 
We then support Task Comprehensiveness by pretraining on varieties of uni-modal and cross-modal tasks.
%, improving the downstream task performance and the generalization ability.

%Then we design tasks instructions for multitask pretraining, which inherently enables the model to transfer to open-domain data and unseen tasks.
%for universal Seq2Seq tasks, .
%, which share an identical input-output schema. 
%We apply a unified Seq2Seq framework with a Transformer encoder-decoder architecture. 
% In this framework, both pretraining and finetuning are essentially sequence-to-sequence learning.
%This architecture supports MA, as both multimodal \& unimodal pretraining and transfer tasks share an input-output schema, and thus there is no need to add task- and modality-specific layers for downstream tasks. 

%To support Task Comprehensiveness, we follow instruction-based training~\cite{flan, t0} is a promising solution, and in principle it does not contradict with MA or TA. Therefore, we design tasks instructions for multitask pretraining, which inherently enables the model to transfer to open-domain data and unseen tasks.

% Furthermore, we design task instructions inherently enabling the model to transfer to open-domain data and unseen tasks.
% Furthermore, with task-specific instructions, the model is capable of transferring to open-domain data and unseen tasks.  
% To encourage the model to better adapt to open-domain data and unseen tasks, we leverage handcrafted task instructions

% Here we should add some more design of the method

To summarize:
\begin{itemize}
    \item We propose \ofa, a Task-Agnostic and Modality-Agnostic framework that supports Task Comprehensiveness. OFA is the first attempt to unify the following vision \& language, vision-only and language-only tasks, including understanding and generation, e.g., text-to-image generation, visual grounding, visual question answering (VQA), image captioning, image classification, language modeling, etc., 
    %and modalities, including multi-modality and uni-modality, 
    via a simple sequence-to-sequence learning framework with a unified instruction-based task representation. 
    \item OFA is pretrained on the publicly available datasets of $20$M image-text pairs, in comparison with recent models that rely on paired data of a much larger scale~\cite{simvlm,florence}. OFA achieves state-of-the-art performances in a series of vision \& language downstream tasks, including image captioning, visual question answering, visual entailment, referring expression comprehension, etc. 
    \item OFA, as a multimodal pretrained model, achieves comparable performances on unimodal tasks with SOTA pretrained models in language or vision, e.g., RoBERTa, ELECTRA and DeBERTa for natural language understanding, UniLM, Pegasus and ProphetNet for natural language generation, and MoCo-v3, BEiT and MAE for image classification.  
    % \item Currently OFA is only trained on a total of 20M public image-text pairs, far less than those used in  previous SOTAs. OFA achieves new SOTAs on multimodal benchmarks, including image captioning (COCO test CIDEr: 149.6), text-to-image generation (COCO test FID: 10.5), VQA (test-std acc.: 80.02), SNLI-VE (test acc.: 90.20), and referring expression comprehension (RefCOCO / RefCOCO+ / RefCOCOg test acc.: 92.93 / 90.10 / 85.20). OFA also performs competitively with uni-modal pretrained models on language and vision tasks, but 
    % outperforms the previous multimodal ones.
    %still most previous multimodal pretrained models far underperform the uni-modal ones. 
    
    \item We verify that \ofa~achieves competitive performance in zero-shot learning. Also, it can transfer to unseen tasks with new task instructions and adapt to out-of-domain information without finetuning. 
    
    % \item \ofa can easily integrate various pretraining tasks through seq2seq learning paradigm to achieve effective knowledge transfer between different tasks. Experimental results demonstrate that \ofa not only outperforms or achieves comparable performance with the state-of-the-art methods on multimodal tasks, but also maintain satisfactory performance in unimodal tasks, including text classification, text generation, and image classification.
    
    % \item We find that \ofa possesses the ability to transfer to unseen tasks with proper instruction, and we show that it is able to achieve competitive performance in the setup of zero-shot learning. 
    % \item We conduct experiments on around 20 tasks across different domains. The experimental results show that \ofa can achieve impressive performance on NLP, CV and VL tasks even without fine-tuning.
\end{itemize}
\section{Related Work}
\label{sec:related-work}
\begin{figure*}[htbp]
\vskip 0.2in
    \centering
    \includegraphics[width=1.\linewidth]{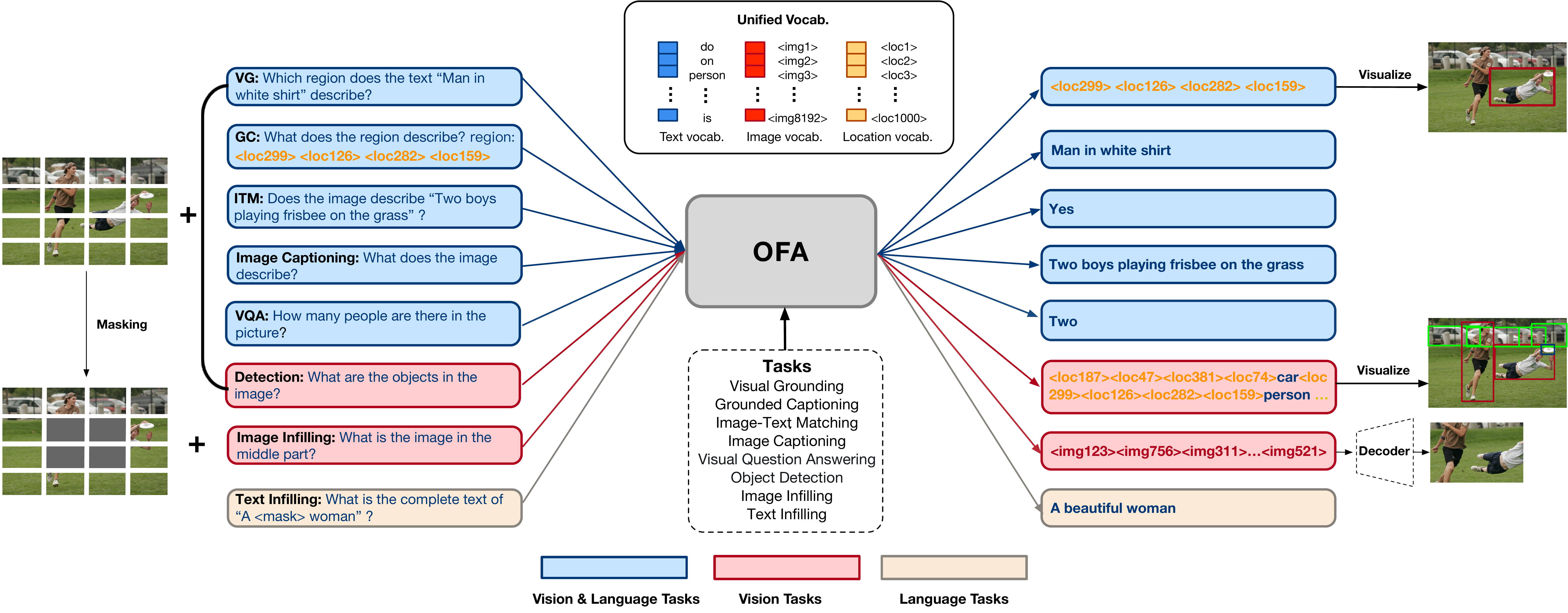}
    \caption{A demonstration of the pretraining tasks, including visual grounding, grounded captioning, image-text matching, image captioning, VQA, object detection, image infilling as well as text infilling.}
    \label{fig:\ofa}
\end{figure*}
% \zc{We need compare to some `unify' works and show the differences, i.e., domain,modality,task agnostic methods like ~\cite{perceiver,perceiverio}, ~\cite{kaiser2017one}, or even older literature. We can organize as 1) LP and VP, 2) SOTA VLP, 3) unify literature as three different setting focusing on different purpose.}
% In this section, we review the previous studies in language pretraining \& vision pretraining, multimodal pretraining, as well as unified frameworks. 

\paragraph{Language Pretraining \& Vision Pretraining} 
Natural language pretraining has revolutionized the whole NLP research community. A representation of this track is the birth of BERT~\cite{bert} and GPT~\cite{gpt}. 
% They demonstrated that pretraining on large-scale data of plain texts with the paradigm of unsupervised learning can provide a superior initialization for models for downstream tasks of language understanding and language generation. 
A number of studies have been progressively advancing pretraining by improving pretraining tasks and designing more sophisticated model architectures~\cite{xlnet, ernie, ernie2.0, roberta, unilm, t5, bart}. 
% Recent advances show that pretrained models at scale can achieve remarkable performance~\cite{t5, gpt3, gshard, switch, ernie3.0}, and \cite{gpt3} and \cite{flan} both show that pretraining on sufficiently large-scale data can endow models with few-shot or zero-shot learning capabilities. 
% With a sophisticated design of manual prompts~\cite{} or learnable soft prompts~\cite{prefix_tuning, prompt_tuning, ptuning, ppt, ptuningv2}, those capabilities can be highly encouraged, and thus we witnessed the upsurge of prompt tuning~\cite{prompt_tuning_survey}. 
% \paragraph{Vision Pretraining} 
Having witnessed the success of natural language pretraining, researchers have promoted self-supervised learning (SSL) in computer vision~\cite{simclr, moco, byol, mocov3}. 
Recently, mirroring masked language modeling (MLM) in language pretraining, generative pretraining~\cite{beit, mae} with ViT architecture~\cite{vit} further boosts downstream performance. 

% Contrastive learning~\cite{cl, infonce} has become a significant method for SSL. Based on the philosophy of alignment and uniformity~\cite{}, 
% A series of works, including SimCLR~\cite{simclr}, MoCo~\cite{moco, mocov3}, BYOL~\cite{byol}, have demonstrated that SSL is capable of promoting model performance in downstream classification tasks. Another trend is the combination of vision transformer and self-supervised learning. MoCo-v3~\cite{mocov3} demonstrated that vision transformer~\cite{vit} can be a promising solution for SSL, and later generative self-supervised pretraining~\cite{beit, mae}, mirroring MLM in language pretraining, leveled up the downstream performance, indicating that VIT plus generative pretraining should be another promising direction for SSL.

\paragraph{Multimodal Pretraining} 
Multimodal pretraining has been developing rapidly~\cite{visualbert, vilbert, vlp, lxmert, uniter, unicoder-vl, interbert, vilbert-mt, e2e-vlp, oscar, villa, vinvl, ernie_vil, unimo, pixelbert}. 
Researchers have applied the masking strategies and the encoder-decoder architecture to adapt models to generation tasks~\cite{oscar, vinvl, m6, simvlm}. 
Besides, to simplify preprocessing, patch projection has received attention and helped Transformer achieve SOTA performance in downstream tasks~\cite{simvlm, vlmo}. 
% These models are mostly pretrained on high-quality datasets of image-text pairs, without leveraging the large-scale weakly supervised ones. 
% The above studies are mostly on high-quality image-text pair data. 
To make full use of large-scale weakly supervised data, \cite{clip} trains a bi-encoder on $400$ million pairs and demonstrates excellent performance in retrieval tasks. 
Another line of work is text-to-image synthesis. A bunch of works~\cite{dalle, cogview, m6, nvwa} incorporate Transformer with VQVAE~\cite{vqvae} or VQGAN~\cite{vqgan} to generate high-quality images with high resolution. 
However, the previously mentioned methods are limited in processing a single type of data, such as cross-modal data only or limited in their capabilities. 
% which denotes that none of them can tackle all cross-modal or unimodal understanding and generation tasks effectively. 
Also, the discrepancy between pretraining and finetuning behaviors limits the transferability to open-ended data. 
\paragraph{Unified Frameworks}
To pursue the unified models, \cite{kaiser2017one} demonstrate a uniform format to represent tasks.
In NLP, recent studies unify diverse tasks covering natural language understanding and generation to text-to-text transfer~\cite{t5} or language modeling~\cite{gpt3}. 
Following this idea, \cite{vlt5} and \cite{unicorn} demonstrate text-generation-based multimodal pretrained models.
\cite{perceiver} and \cite{perceiverio} propose a simple framework that can process information from multiple modalities with a uniform byte-sequence representation.
\cite{unit} and \cite{flava} unify tasks of different modalities by designing various task-specific layers. 
\cite{uni-perceiver} explores to employ a retrieval-based unified paradigm.
% A series of studies try to explore unification for modalties and tasks~\cite{vlt5, perceiver, unit, flava, unicorn, uni-perceiver}. 
However, these multimodal pretrained models suffer from performance degradation in downstream tasks, e.g., VQA, image captioning, etc., and they have no image generation capability.

\section{\ofa}
\label{sec:\ofa}

In this work, we propose \ofa, a unified Seq2Seq framework for the unification of I/O \& architectures, tasks, and modalities. The 
overall framework is illustrated in Figure~\ref{fig:\ofa}.

\subsection{I/O \& Architecture}
\label{sec:unified_io_arc}
% \zc{The title of this subsection need change. We don't have any contribution in architecture in fact. The focus should be anything like task-agnostic IO, or unified multimodal IO.}

\paragraph{I/O} The most common practice of multimodal pretraining is the pretraining of Transformer models on image-text pair corpus at scale. 
This requires data preprocessing or modality-specific adaptors to enable the joint training of both visual and linguistic information with the Transformer architecture. 
% To enable multimodal pretraining, it is necessary to preprocess the data so that both visual and linguistic information can be jointly processed by the Transformer. 
% The conventional method for processing images is the extraction of object features with an object detector, e.g. Faster RCNN~\cite{faster-rcnn}. However, although effective, they are complicated and time-consuming especially for the inference~\cite{vilt}. 
Compared with the complex, resource\&time-consuming object feature extraction, we aim for simplicity and directly use ResNet modules to convolve $\textrm{x}_v \in \mathbb{R}^{H \times W \times C}$ to $P$ patch features of the hidden size, following \cite{coatnet} and \cite{simvlm}. 
% The output visual feature sequence is $\textrm{h}_v \in \mathbb{R}^{L_v \times H}$, where $L_v = \frac{HW}{P^2}$. Simple linear projection has advantages in efficiency, but it still suffers from performance downgrade in comparison with convolution~\cite{convit, early}. 
% Following \cite{coatnet} and \cite{simvlm}, we add convolution layers before Transformer layers for contextualized patch representations. Specifically, we use the first three blocks of ResNet~\cite{resnet} to reach this goal. 
As to processing the linguistic information, we follow the practice of GPT~\cite{gpt} and BART~\cite{bart} that we apply byte-pair encoding (BPE)~\cite{bpe} to the given text sequence to transform it into a subword sequence and then embed them to features. 

To process different modalities without task-specific output schema, it is essential to represent data of various modalities in a unified space. 
A possible solution is to discretize text, image, and object and represent them with tokens in a unified vocabulary. 
% including the vocabulary space of texts, the discrete code space of images and the bucketized coordinates space of bounding boxes.
% Another focus of this work is how to process images for decoding.
% Learning to generate images is one of the pretraining tasks, and it is hard for models to predict continuous representations. 
Recent advances in image quantization~\cite{vqvae, vqgan} have demonstrated effectiveness in text-to-image synthesis~\cite{dalle, m6, cogview, ufc}, and thus we utilize this strategy for the target-side image representations. Sparse coding is effective in reducing the sequence length of image representation. For example, an image of the resolution of $256 \times 256$ is represented as a code sequence of the length of $16 \times 16$. Each discrete code strongly correlates with the corresponding patch~\cite{beit}. 

Apart from representing images, it is also essential to represent objects within images as there are a series of region-related tasks. 
Following \cite{pix2seq}, we represent objects as a sequence of discrete tokens. To be more specific, for each object, we extract its label and its bounding box.
% (the coordinates of the top-left and bottom-right corner, namely $\langle x_1, y_1, x_2, y_2 \rangle$, where $x$ and $y$ respectively represents the horizontal and vertical coordinates). 
The continuous corner coordinates (the top left and the bottom right) of the bounding box are uniformly discretized to integers as location tokens $\langle x_1, y_1, x_2, y_2 \rangle$. 
As to the object labels, they are intrisically words and thus can be represented with BPE tokens. 
% We split the height and width into bins and bucketize the continuous coordinates into discrete location ids. 
% For example, as shown in Figure~\ref{fig:\ofa}, a person on the image can be represented as $``\langle loc187 \rangle, \langle loc47 \rangle, \langle loc381 \rangle, \langle loc74 \rangle, person"$, in which $person$ is the object label represented with BPE tokenization. 
% As there is no clarified sequential relations between objects, we concatenate object sequences in a descending order based on area. 
% We concatenate the objects into a sequence by descending order based on object area.
% In comparison with random ordering, it is easier for \ofa to model such relation.
% The decoder learns to generate these tokens autoregressively.

Finally, we use a unified vocabulary for all the linguistic and visual tokens, including subwords, image codes, and location tokens. 

\paragraph{Architecture}
Following the previous successful practices in multimodal pretraining~\cite{uniter, vinvl, simvlm}, 
% , natural language pretraining~\cite{bert, gpt}, and vision pretraining~\cite{mae, beit, vit}, 
we choose Transformer as the backbone architecture, and we adopt the encoder-decoder framework as the unified architecture for all the pretraining, finetuning, and zero-shot tasks.
Specifically, both the encoder and the decoder are stacks of Transformer layers. A Transformer encoder layer consists of a self attention and a feed-forward network (FFN), while a Transformer decoder layer consists of a self attention, an FFN and a cross attention for building the connection between the decoder and the encoder output representations. To stabilize training and accelerate convergence, we add head scaling to self attention, a post-attention layer normalization (LN)~\cite{layer_norm}, and an LN following the first layer of FFN~\cite{normformer}. For positional information, we use two absolute position embeddings for text and images, respectively. Instead of simply adding the position embeddings, we decoupling the position correlation from token embeddings and patch embeddings~\cite{tupe}. In addition, we also use 1D relative position bias for text~\cite{t5} and 2D relative position bias for image~\cite{simvlm,coatnet}.

% Note that pretraining Transformer models at scale may suffer from training instabilities~\cite{normformer, cogview}. To alleviate the hassles, we apply methods in NormFormer~\cite{normformer} to the Transformer layer to stabilize the training process and accelerate convergence. To be more specific, we add head scaling to self attention, and add a post-attention layer normalization (LN)~\cite{layer_norm} and an LN after the first layer of FFN.

\subsection{Tasks \& Modalities}
A unified framework is designed to provide architecture compatibility across different modalities and downstream tasks so that opportunities can arise to generalize to unseen tasks within the same model. Then we have to represent the possible downstream tasks concerning different modalities in a unified paradigm.
% \zc{There seems  to have duplications in 3.1, 3.2 , 3.3 and experiments.}
% A fundamental objective for foundation models is the unification of tasks and modalities. A foundation model should be able to robustly adapt to different types of downstream tasks, e.g., understanding, generation, etc., concerning different modalities, including uni-modal tasks, e.g. text classification, text similarity, abstractive summarization~\cite{rush}, image classification~\cite{imagenet}, object detection, etc., and cross-modal tassk, including visual question answering~\cite{vqa}, image captioning~\cite{coco_cap}, referring objects~\cite{refcoco}, etc. 
Therefore, an essential point for the design of pretraining tasks is the consideration of multitask and multimodality.

% As mentioned in Sec.~\ref{sec:unified_io_arc}, we unify the formats of inputs and outputs and we also demonstrate a unified architecture. 
To unify tasks and modalities, we design a unified sequence-to-sequence learning paradigm for pretraining, finetuning, and inference on all tasks concerning different modalities. 
Both pretraining tasks and downstream tasks of cross-modal and uni-modal understanding and generation are all formed as Seq2Seq generation. 
It is available to perform multitask pretraining on multimodal and uni-modal data to endow the model with comprehensive capabilities. 
Specifically, we share the identical schema across all tasks, while we specify handcrafted instructions for discrimination~\cite{flan}. 
% During pretraining, we unify all tasks to the form of Seq2Seq learning and optimizing the entire
% unified vocabulary. 
% Furthermore, we also adopt this paradigm for our finetuning and inference. All tasks including generation and classification are formed as Seq2Seq generation tasks. 

% In order to achieve this objective, we construct our pretraining dataset by incorporating cross-modal data, i.e., image-text pairs, and uni-modal data, i.e., plain texts and images. Specifically, the cross-modal data includes three types of data, namely image-caption data, image-question-answer data, as well as object-description data.
% The plain text data includes sentences or passages from the commonly used datasets for natural language pretraining. 
% The image data includes raw images from multiple datasets, and we exclude annotations or labels for unsupervised pretraining. 
% We refer more details about datasets to Sec.\ref{sec:datasets} and Appendix~\ref{sec:appendix_pretraining}. 
% We unify the pretraining task with the form of Seq2Seq learning, and thus it is available to design multiple tasks to endow the model with sufficient capabilities in cross-modal and uni-modal representation learning by only specifying the formats of inputs and outputs and handcrafting manual task instructions for discrimination, following \cite{flan}. 

For cross-modal representation learning, we design $5$ tasks, including visual grounding (VG), grounded captioning (GC), image-text matching (ITM), image captioning (IC), and visual question answering (VQA). 
% For VG, we use the data in Visual Genome Captions~\cite{vg}, where each caption describe a specific region. 
For VG, the model learns to generate location tokens specifying the region position $\langle x_1, y_1, x_2, y_2 \rangle$ based on the input of the image $x^i$ and the instruction ``Which region does the text $x^t$ describe?'' where $x^t$ refers to the region caption.
GC is an inverse task of VG. The model learns to generate a description based on the input image $x^i$ and the instruction ``What does the region describe? region: $\langle x_1, y_1, x_2, y_2 \rangle$''. 
For ITM, we use each original image-text pair as the positive sample and construct a new one as the negative by pairing the image with a randomly substituted caption. 
The model learns to discriminate whether the given image and text are paired by learning to generate ``Yes'' or ``No'' based on the input image $x^i$ and the instruction ``Does the image describe $x^t$?''. 
As to image captioning, this task can naturally adapt to the sequence-to-sequence format. The model learns to generate the caption based on the given image and the instruction ``What does the image describe?''. 
For VQA, we send the image and the question as the input and require the model to learn to generate correct answers. 

For uni-modal representation learning, we design $2$ tasks for vision and $1$ task for language, respectively. The model is pretrained with image infilling and object detection for vision representation learning. Recent advances in generative self-supervised learning for computer vision show that masked image modeling is an effective pretraining task~\cite{beit, mae}. 
In practice, we mask the middle part of the images as the input. The model learns to generate the sparse codes for the central part of the image based on the corrupted input and the specified instruction ``What is the image in the middle part?''. 
We additionally add object detection to pretraining following~\cite{e2e-vlp}. The model learns to generate human-annotated object representations, i.e., the sequence of object position and label, based on the input image and the text ``What are the objects in the image?'' as the instruction. 
Both tasks strengthen the representation learning on both pixel and object levels. 
For language representation learning, following the practice of \cite{bart}, we pretrain the unified model on plain text data with text infilling.

In this way, we unify multiple modalities and multiple tasks to a single model and pretraining paradigm. \ofa~is pretrained jointly with those tasks and data. Thus, it can perform different tasks concerning natural language, vision, and cross-modality. 

\subsection{Pretraining Datasets}
\label{sec:datasets}
We construct pretraining datasets by incorporating Vision \& Language data (i.e., image-text pairs), Vision data (i.e., raw image data, object-labeled data), and Language data (i.e., plain texts). 
For replication, we only use datasets that are publicly available. 
We carefully filter our pretraining data and exclude images that appear in the validation and test sets of downstream tasks to avoid data leakage. 
We provide more details about pretraining datasets in Appendix~\ref{sec:appendix_pretraining_datasets}.

\subsection{Training \& Inference}
\label{sec:finetune_inference}

We optimize the model with the cross-entropy loss. Given an input $x$, an instruction $s$ and an output $y$, we train \ofa~by minimizing $\mathcal{L} = -\sum_{i=1}^{|y|}{\rm log}P_{\theta}(y_i|y_{<i},x,s)$,
where $\theta$ refers to the model parameters. 
For inference, we apply the decoding strategies, e.g., beam search, to enhance the quality of generation.
However, this paradigm has several problems in classification tasks:
1. optimizing on the entire vocabulary is unnecessary and inefficient;
2. the model may generate invalid labels out of the closed label set during inference.
To overcome these issues, we introduce a search strategy based on prefix tree (Trie, \cite{trie}). Experimental results show that the Trie-based search can enhance the performance of \ofa~on classification tasks. See Appendix~\ref{sec:trie_search} for more details.

\subsection{Scaling Models}
% In order to investigate how OFA of different model sizes perform in downstream tasks, we develop a series of OFA models. First and foremost, we build $\rm \ofa_{Base}$ and $\text{OFA}_{Large}$, whose model sizes are generally comparable to $\text{SimVLM}_{Base}$ and $\text{SimVLM}_{Large}$. To investigate the effects of model scaling, we develop a \textit{Huge}-size model $\text{OFA}_{Huge}$ in comparison with $\text{SimVLM}_{Huge}$. Additionally, based on the assumption that unified pretraining and large-scale data corpus can also benefit smaller models, we develop two small models, $\text{OFA}_{Medium}$ and $\text{OFA}_{Tiny}$. The size of $\text{OFA}_{Medium}$ is around half smaller than that of $\text{OFA}_{Base}$, and the size of $\text{OFA}_{Tiny}$ is less than $20\%$ of it. 
% For more details of model implementation, we list the detailed hyperparameters in Appendix~\ref{sec:appendix_pretrain_details}. 

\begin{table}[t]
\caption{Detailed hyperparameters of OFA model configuration. We list the configuration for OFA of $5$ different sizes.
}
\vskip 0.15in
\centering
\begin{adjustbox}{max width=1.\textwidth}
\begin{tabular}{lccccccc}
\toprule
  Model
  &\#Param.
  &Backbone
  &Hidden size
  &Intermediate Size
  &\#Head
  &\#Enc. Layers
  &\#Dec. Layers
  \\
\midrule
  $\text{OFA}\rm_{Tiny}$
  &33M
  &ResNet50
  &256
  &1024
  &4
  &4
  &4
  \\
  $\text{OFA}\rm_{Medium}$
  &93M
  &ResNet101
  &512
  &2048
  &8
  &4
  &4
  \\
  $\text{OFA}\rm_{Base}$
  &182M
  &ResNet101
  &768
  &3072
  &12
  &6
  &6
  \\
  $\text{OFA}\rm_{Large}$
  &472M
  &ResNet152
  &1024
  &4096
  &16
  &12
  &12
  \\
  $\text{OFA}\rm_{Huge}$
  &930M
  &ResNet152
  &1280
  &5120
  &16
  &24
  &12
  \\
\bottomrule
\end{tabular}
\end{adjustbox}
\label{tb:model_configuration}
\end{table}
In order to investigate how OFA of different model sizes perform in downstream tasks, we have developed $5$ versions of OFA models, scaling from $33$M to $940$M parameters, and we list their detailed hyperparameters in Table~\ref{tb:model_configuration}. 

To be more specific, we have built basic models of $\rm Base$ and $\rm Large$ sizes, $\text{OFA}\rm_{Base}$ and $\text{OFA}\rm_{Large}$. As our network configuration is similar to BART~\cite{bart}, their sizes are similar to those of $\text{BART}\rm_{Base}$ and $\text{BART}\rm_{Large}$. 
Additionally, we have developed OFA of a larger size, which we name it $\text{OFA}\rm_{Huge}$, or OFA without specific mentioning in the tables. Its size is comparable to that of $\text{SimVLM}\rm_{Huge}$ or $\text{ViT}\rm_{Huge}$. To investigate whether smaller OFA can still reach satisfactory performance, we have developed $\text{OFA}\rm_{Medium}$ and $\text{OFA}\rm_{Tiny}$, which are solely around half and less than $20\%$ as large as $\text{OFA}\rm_{Base}$.

\section{Experiments}
\label{sec:experiments}
This section provides experimental details and analyses to demonstrate our model's effectiveness. 
See Appendix~\ref{app:implementation_details} for implementation details.

\subsection{Results on Cross-modal Tasks}
\begin{table*}[t]
% \caption{Experimental results on cross-modal downstream tasks. \ofa~can reach the SOTA performance in both image captioning and referring expression comprehension. On classification tasks, i.e. VQA and SNLI-VE, our generation model can reach competitive performance with the concurrent SOTA classification models. All reported results are from \textit{Large} models, which have similar amount of parameters. }
\caption{Experimental results on cross-modal understanding tasks including VQA and visual entailment. Note that we report the best results from the previous SOTAs, and specifically SimVLM is a huge-size model comparable to ViT-Huge pretrained on 1.8B image-text pairs, and Florence is built with CoSwin-H and $\text{RoBERTa}$ and it is pretrained on 900M image-text pairs.}
\center
\vskip 0.15in
\begin{adjustbox}{max width=1.\textwidth}
\begin{tabular}{@{\extracolsep{\fill}}lcccc}
\toprule
  \multirow{2}*{Model}
%   &COCO Captions
  & \multicolumn{2}{c}{VQA}
  & \multicolumn{2}{c}{SNLI-VE}
%   &RefCOCO
%   &RefCOCO+
%   &{RefCOCOg}
  \\
%   &{B@4 / M / C / S}
  & test-dev & test-std
  & dev & test
%   &{val / testA / testB}
%   &{val / testA / testB}
%   &{val-u / test-u}
  \\
\midrule
%   VisualBERT \cite{visualbert}
%   &-
%   &70.80 / 71.00
%   &-
%   &-
%   &-
%   &-
%   \\
%   ViLBERT \cite{vilbert}
% %   &-
%   &70.55 & 70.92
%   &- & -
% %   &-
% %   &72.34 / 78.52 / 62.61
% %   &-
%   \\
%   VL-BERT \cite{vlbert}
% %   &-
%   &71.79 / 72.22
%   &- & -
% %   &-
% %   &72.59 / 78.57 / 62.30
% %   &-
%   \\
%   LXMERT \cite{lxmert}
%   &-
%   &72.42 / 72.54
%   &-
%   &-
%   &-
%   &-
%   \\
  UNITER \cite{uniter}
%   &-
%   &73.82 & 74.02
  & 73.8 & 74.0
%   &79.39 & 79.38
  & 79.4 & 79.4
%   &81.41 / 87.04 / 74.17
%   &75.90 / 81.45 / 66.70
%   &74.86 / 75.77
  \\
  OSCAR \cite{oscar}
%   &41.7 / 30.6 / 140.0 / 24.5
%   &73.61 & 73.82
  &73.6 & 73.8
  &- & -
%   &-
%   &-
%   &-
  \\
%   Pixel-BERT \cite{pixelbert}
%   &-
%   &74.45 / 74.55
%   &-
%   &-
%   &-
%   &-
%   \\
  VILLA \cite{villa}
%   &-
%   & 74.69 & 74.87
  & 74.7 & 74.9
%   & 80.18 & 80.02
  &80.2 & 80.0
%   &82.39 / 87.48 / 74.84
%   &76.17 / 81.54 / 66.84
%   &76.18 / 76.71
  \\
  VL-T5 \cite{vlt5}
%   &34.5 / 28.7 / 116.5 / 21.9
  & - & 70.3
  &- & -
%   &-
%   &71.3 / - / -
%   &-
  \\
%   MDETR \cite{Kamath2021MDETRM}
% %   &-
%   &70.64 / 70.63
%   &-
%   &86.75 / 89.58 / 81.41
%   &79.52 / 84.09 / 70.62
%   &81.64 / 80.89
%   \\
%   UNICORN \cite{unicorn}
% %   &35.8 / 28.4 / 119.1 / 21.5
%   &69.2 / 69.4
%   &-
%   &88.29 / 90.42 / 83.06
%   &80.30 / 85.05 / 71.88
%   &83.44 / 83.93
%   \\
%   SOHO \cite{soho}
%   &-
%   &73.25 / 73.47
% %   &\textcolor{blue}{85.00} / \textcolor{blue}{84.95}
%   &85.00 / 84.95
%   &-
%   &-
%   &-
%   \\
%   KD-VLP \cite{Liu2021KDVLPIE}
%   &-
%   &74.20 / 74.31
% %   &78.21(\textcolor{blue}{88.18}) / 77.87(\textcolor{blue}{88.21})
%   & 88.18 / 88.21
%   &-
%   &-
%   &-
%   \\
  VinVL \cite{vinvl}
%   &41.0 / 31.1 / 140.9 / 25.2
%   &76.52 & 76.60
  &76.5 & 76.6
  &- & -
%   &-
%   &-
%   &-
  \\
%   ViLT \cite{vilt}
%   &-
%   &70.94 / -
%   &-
%   &-
%   &-
%   &-
%   \\
%   ALBEF \cite{albef}
%   &-
%   &75.84 / 76.04
%   &80.80 / 80.91
%   &-
%   &-
%   &-
%   \\
  UNIMO \cite{unimo}
%   &39.6 / - / 127.7 / -
%   &75.06 & 75.27
  &75.0 & 75.3
%   &81.11 & 80.63
  &81.1 & 80.6
%   &-
%   &-
%   &-
  \\
  ALBEF \cite{albef}
  &75.8 & 76.0
  &80.8 & 80.9
  \\
%   X-VLM \cite{Zeng2021MultiGrainedVL}
%   &-
%   &76.77 / 76.89
%   &-
%   &-
%   &76.31 / 82.78 / 67.40
%   &-
%   \\
  METER \cite{meter}
%   &-
%   &77.68 & 77.64
  &77.7 & 77.6
%   &80.86 & 81.19
  &80.9 & 81.2
%   &-
%   &-
%   &-
  \\
  VLMo \cite{vlmo}
%   &-
%   &79.94 & 79.98
  &79.9 & 80.0
  &- & -
%   &-
%   &-
%   &-
  \\
  SimVLM \cite{simvlm}
%   &40.3 / \textbf{33.4} / 142.6 / 24.7
%   &79.32 & 79.56
  &80.0 & 80.3
  &86.2 & 86.3
%   &-
%   &-
%   &-
  \\
  Florence \cite{florence}
  &80.2 & 80.4
  &- & -
  \\
\midrule
%   \ofa
%   &\textbf{43.5} / 31.9 / \textbf{149.6} / \textbf{26.1}
%   &79.87 / \textbf{80.02}
% %   &81.42(\textcolor{blue}{90.30}) / 81.61(\textcolor{blue}{90.00})
%   & \textbf{90.30} / \textbf{90.20} 
%   &\textbf{90.05} / \textbf{92.93} / \textbf{85.26}
%   &\textbf{84.49} / \textbf{90.10} / \textbf{77.77}
%   &\textbf{84.54} / \textbf{85.20}
%   \\
$\text{OFA}\rm_{Tiny}$
%   & - / - / \textbf{154.6} / -
%   & 77.98 & 78.07
    &70.3 & 70.4
%   &81.42(\textcolor{blue}{90.30}) / 81.61(\textcolor{blue}{90.00})
  & 85.3 & 85.2
  \\
  $\text{OFA}\rm_{Medium}$
%   & - / - / \textbf{154.6} / -
%   & 77.98 & 78.07
    &75.4 & 75.5
%   &81.42(\textcolor{blue}{90.30}) / 81.61(\textcolor{blue}{90.00})
  & 86.6 & 87.0
  \\
  $\text{OFA}\rm_{Base}$
%   & - / - / \textbf{154.6} / -
%   & 77.98 & 78.07
    &78.0 & 78.1
%   &81.42(\textcolor{blue}{90.30}) / 81.61(\textcolor{blue}{90.00})
  & 89.3 & 89.2
%   & 88.48 / 90.67 / 83.30
%   & 81.39 / 87.15 / 74.29
%   &82.29 / 82.31
  \\
  $\text{OFA}\rm_{Large}$
%   & - / - / \textbf{154.6} / -
%   & 80.34 & 80.45
% result w/o EL
  &80.3 & 80.5
%   &81.42(\textcolor{blue}{90.30}) / 81.61(\textcolor{blue}{90.00})
  & 90.3 & 90.2
%   & 90.05 / 92.93 / 85.26
%   & 85.80  / 89.87 / 79.22
%   &85.89 / 86.55
  \\
  $\text{OFA}$
%   & - / - / \textbf{154.6} / -
  &\textbf{82.0} & \textbf{82.0}
%   &81.42(\textcolor{blue}{90.30}) / 81.61(\textcolor{blue}{90.00})
  & \textbf{91.0} & \textbf{91.2} 
%   &\textbf{91.62} / \textbf{93.87} / \textbf{87.60}
%   &\textbf{87.29} / \textbf{91.65} / \textbf{80.24}
%   &\textbf{88.15} / \textbf{88.13}
  \\
% \ \ \textit{w/o finetuning}
%   &25.7 / 22.1 / 84.0 / 16.7
%   &72.30 / 72.23
%   &-
%   &76.43 / 81.60 / 71.85
%   &65.79 / 73.56 / 56.49
%   &68.65 / 69.36
%   \\
\bottomrule
\end{tabular}
\end{adjustbox}
\label{tb:vqa-ve}
\end{table*}

\begin{table*}[t]
\caption{Experimental results on MSCOCO Image Captioning. We report the results on the Karpathy test split. Note that SimVLM and LEMON are huge-size models. }
\center
\vskip 0.15in
\begin{adjustbox}{max width=1.\textwidth}
\begin{tabular}{@{\extracolsep{\fill}}lcccccccc}
\toprule
  \multirow{2}*{Model}
  &\multicolumn{4}{c}{Cross-Entropy Optimization}
  &\multicolumn{4}{c}{CIDEr Optimization}
  \\
  &BLEU@4  & METEOR & CIDEr & SPICE
  &BLEU@4  & METEOR & CIDEr & SPICE
  \\
\midrule
  VL-T5 \cite{vlt5}
  &34.5 & 28.7 & 116.5 & 21.9
  &- & - & - & -
  \\
  OSCAR \cite{oscar}
  &37.4 & 30.7 & 127.8 & 23.5
  &41.7 & 30.6 & 140.0 & 24.5
  \\
  UNICORN \cite{unicorn}
  &35.8 & 28.4 & 119.1 & 21.5
  &- & - &- & -
  \\
  VinVL \cite{vinvl}
  &38.5 & 30.4 & 130.8 & 23.4
  &41.0 & 31.1 & 140.9 & 25.2
  \\
  UNIMO \cite{unimo}
  &39.6 &  - & 127.7 & -
  &- & - & - & -
  \\
  LEMON \cite{lemon}
  &41.5 &  30.8 & 139.1 & 24.1
  &42.6 & 31.4 & 145.5 & 25.5
  \\
  SimVLM \cite{simvlm}
  &40.6 & \textbf{33.7} & 143.3 & \textbf{25.4}
  &- & - & - & -
  \\
\midrule
%   \ofa
%   &\textbf{43.5} / 31.9 / \textbf{149.6} / \textbf{26.1}
%   &79.87 / \textbf{80.02}
% %   &81.42(\textcolor{blue}{90.30}) / 81.61(\textcolor{blue}{90.00})
%   & \textbf{90.30} / \textbf{90.20} 
%   &\textbf{90.05} / \textbf{92.93} / \textbf{85.26}
%   &\textbf{84.49} / \textbf{90.10} / \textbf{77.77}
%   &\textbf{84.54} / \textbf{85.20}
%   \\
$\text{OFA}\rm_{Tiny}$
  & 35.9 & 28.1 & 119.0 & 21.6
  & 38.1 & 29.2 & 128.7 & 23.1
  \\
 $\text{OFA}\rm_{Medium}$
  & 39.1 & 30.0 & 130.4 & 23.2
  & 41.4 & 30.8 & 140.7 & 24.8
  \\
 $\text{OFA}\rm_{Base}$
  & 41.0 & 30.9 & 138.2 & 24.2
  & 42.8 & 31.7 & 146.7 & 25.8
  \\
  $\text{OFA}\rm_{Large}$
  & 42.4 & 31.5 & 142.2 & 24.5
  & 43.6 & 32.2 & 150.7 & 26.2
  \\
  $\text{OFA}$
  & \textbf{43.9} & 31.8 & \textbf{145.3} & 24.8
  & \textbf{44.9} & \textbf{32.5} & \textbf{154.9} & \textbf{26.6}
  \\
% \ \ \textit{w/o finetuning}
%   &25.7 / 22.1 / 84.0 / 16.7
%   &72.30 / 72.23
%   &-
%   &76.43 / 81.60 / 71.85
%   &65.79 / 73.56 / 56.49
%   &68.65 / 69.36
%   \\
\bottomrule
\end{tabular}
\end{adjustbox}
\label{tb:caption}
\end{table*}

\begin{table*}[t]
\caption{Experimental results on the $3$ datasets of referring expression comprehension, namely RefCOCO, RefCOCO+, and RefCOCOg. We report the Acc@0.5 on different test splits of the datasets. }
\center
\vskip 0.15in
\begin{adjustbox}{max width=1.\textwidth}
\begin{tabular}{@{\extracolsep{\fill}}lcccccccc}
\toprule
  \multirow{2}*{Model}
%   &COCO Captions
  &\multicolumn{3}{c}{RefCOCO}
  &\multicolumn{3}{c}{RefCOCO+}
  &\multicolumn{2}{c}{RefCOCOg}
  \\
  & val & testA & testB
  & val & testA & testB
  &val-u & test-u
  \\
\midrule
  VL-T5 \cite{vlt5}
  &- & - & -
  &- & - & -
  &- & 71.3
  \\
  UNITER \cite{uniter}
  &81.41 & 87.04 & 74.17
  &75.90 & 81.45 & 66.70
  &74.86 & 75.77
  \\
  VILLA \cite{villa}
  &82.39 & 87.48 & 74.84
  &76.17 & 81.54 & 66.84
  &76.18 & 76.71
  \\
  MDETR \cite{Kamath2021MDETRM}
  &86.75 & 89.58 & 81.41
  &79.52 & 84.09 & 70.62
  &81.64 & 80.89
  \\
  UNICORN \cite{unicorn}
  &88.29 & 90.42 & 83.06
  &80.30 & 85.05 & 71.88
  &83.44 & 83.93
  \\
\midrule
  $\text{OFA}\rm_{Tiny}$
  & 80.20 & 84.07 & 75.00
  & 68.22 & 75.13 & 57.66
  &72.02 & 69.74
  \\
  $\text{OFA}\rm_{Medium}$
  & 85.34 & 87.68 & 77.92
  & 76.09  & 83.04 & 66.25
  &78.76 & 78.58
  \\
  $\text{OFA}\rm_{Base}$
  & 88.48 & 90.67 & 83.30
  & 81.39 & 87.15 & 74.29
  &82.29 & 82.31
  \\
  $\text{OFA}\rm_{Large}$
  & 90.05 & 92.93 & 85.26
  & 85.80  & 89.87 & 79.22
  &85.89 & 86.55
  \\
  \ofa
  &\textbf{92.04} & \textbf{94.03} & \textbf{88.44}
  &\textbf{87.86} & \textbf{91.70} & \textbf{80.71}
  &\textbf{88.07} & \textbf{88.78}
  \\
\bottomrule
\end{tabular}
\end{adjustbox}
\label{tb:refcoco}
\end{table*}

We evaluate our models on different cross-modal downstream tasks, covering cross-modal understanding and generation. Specifically, we implement experiments on multimodal understanding datasets including VQAv2 for visual question answering and SNLI-VE~\cite{snli-ve} for visual entailment, and multimodal generation including MSCOCO Image Caption~\cite{coco_cap} for image captioning, RefCOCO / RefCOCO+ / RefCOCOg~\cite{refcoco,refcocog} for referring expression comprehension as this task can be viewed as bounding box generation, and MSCOCO Image Caption for text-to-image generation. More details are provided in Appendix~\ref{sec:appendix_downstream}.

% Cross-modal tasks include image captioning on MS COCO Caption~\cite{coco_cap}, visual question answering on VQAv2~\cite{vqav2}, visual entailment on SNLI-VE~\cite{snli-ve}, referring expression comprehension on RefCOCO / RefCOCO+ / RefCOCOg~\cite{refcoco,refcocog}, and text-to-image generation on MS COCO Caption. 

% Note that we use images of the resolution of $640 \times 640$ for $\text{OFA}\rm_{Large}$ and $\text{OFA}\rm_{Huge}$, and in the other cases we all use image of the resolution of $480 \times 480$. 
% Furthermore, we compare beam search with all candidate evaluation (i.e., generating a score for each candidate in the teacher-forcing mode), and find that the latter steadily outperforms the former with a slight advantage, and thus report the results of all candidate evaluation. 
% To alleviate the resolution difference at the pretraining and finetuning stages, we interpolate the positional embeddings with bilinear interpolation following \cite{vit}, and we find that this method can outperform building a new positional embedding for finetuning. 

Table~\ref{tb:vqa-ve} presents the performance of \ofa~and baseline models on VQA and SNLI-VE. 
In general, OFA achieves the best performance in both tasks with $82.0$ on the VQA test-std set and $91.2$ on the SNLI-VE test set. For smaller-size models, $\text{OFA}\rm_{Large}$ can outperform the recent SOTAs, e.g., VLMo and SimVLM, and $\text{OFA}\rm_{Base}$ can beat the SOTAs before the aforementioned two models in both tasks. This demonstrates that OFA can achieve superior performance on cross-modal understanding tasks and scaling up OFA can bring significant improvements, reflecting the strong potential of large-scale pretrained models.

Table~\ref{tb:caption} presents the performance of \ofa~and baseline models on the MSCOCO image captioning dataset. 
We report the results on the Karpathy test split, and we demonstrate the performance of models trained with Cross-Entropy optimization and additionally with CIDEr optimization based on reinforcement learning. 
In comparison with the previous SOTA $\text{SimVLM}\rm_{Huge}$ for Cross-Entropy optimization, OFA outperforms it by around $2$ points in CIDEr evaluation. For CIDEr optimization, OFA of the $3$ sizes all outperform the huge-size LEMON, and OFA demonstrates a new SOTA of $154.9$ CIDEr score. 
By May 31 2022, the single-model OFA had topped the MSCOCO Image Caption Leaderboard.\footnote{\url{https://competitions.codalab.org/competitions/3221\#results}}

% Specifically, on image captioning, \ofa~performs the best on CIDEr evaluation (CIDEr: 149.6). It also surpasses SimVLM, which uses $1.8$ billion image-text pairs for pretraining (around $75\times$ larger than ours), by a large margin of $7.0$.
% At the time of submitting this paper, \ofa~also achieves No.1 on the COCO image captioning online leaderboard.\footnote{\url{https://competitions.codalab.org/competitions/3221\#results}}
To evaluate the capability of visual grounding, we conduct experiments on RefCOCO, RefCOCO+, and RefCOCOg. 
While we unify locations to the vocabulary, visual grounding can be viewed as a sequence generation task. As there is only one target for each query, we limit the generation length to $4$ in order to generate a bounding box by $<x_1, y_1, x_2, y_2>$. 
Experimental results in Table~\ref{tb:refcoco} show that \ofa~reaches the SOTA performance on the $3$ datasets. 
Compared with the previous SOTA UNICORN~\cite{unicorn}, \ofa~achieves significant improvement with a gain of $3.61$, $6.65$ and $4.85$ points on the testA sets of RefCOCO and RefCOCO+ as well as the test-u set of RefCOCOg.

\begin{figure*}[t]
    \centering
    \includegraphics[width=1.\linewidth]{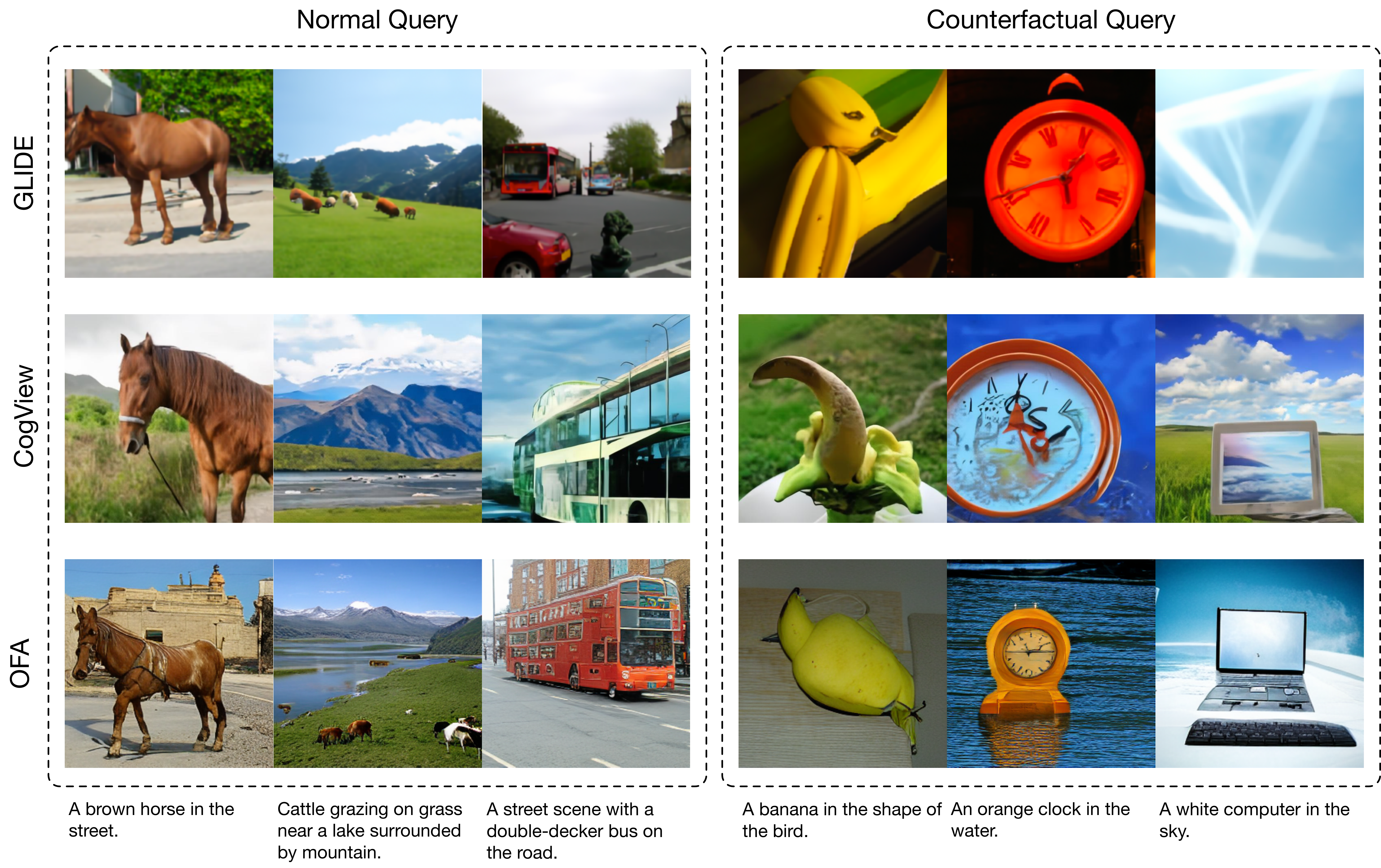}
    \caption{Qualitative comparison with state-of-the-art models for text-to-image generation task. We present more qualitative examples of text-to-image generation for better demonstration in Appendix~\ref{sec:qualitative_examples}.}
    \label{fig:image_gen_samples}
\end{figure*}
% which demonstrates that our method has obvious advantages in recognizing objects in more categories and understanding semantic descriptions like appearance.
% On VQA, \ofa~achieves $80.02$ on test-std and outperforms the SOTA models of the \textit{Large} size, including SimVLM and VLMo. 
% Note that both the previous SOTAs are classification-based models.
% On SNLI-VE, \ofa~achieves state-of-the-art performance and outperforms previous models by a large margin, demonstrating its capability of complex visual-linguistic reasoning.
% On SNLI-VE, \ofa~can achieve similar performance with UNIMO, and it can outperform all the baselines, including SOHO and KD-VLP, in the setup of caption-enhanced SNLI-VE. 

Text-to-image generation is a challenging task even for pretrained model. As we pretrain OFA with the task ``image-infilling'', i.e., recovering masked patches by generating the corresponding codes~\cite{beit}, and thus OFA is able to generate code. We thus directly finetune OFA on the MSCOCO Image Caption dataset for text-to-code generation. At the inference stage, we additionally transform the generated codes to an image with the code decoder. 
Specifically, we use the codes from VQGAN~\cite{vqgan} following \cite{nvwa}. 
% Table~\ref{tb:image-gen-results} demonstrates the model performance of $\text{OFA}\rm_{Base}$ and $\text{OFA}\rm_{Large}$ as well as the baselines on MSCOCO test set.
Experimental results show that \ofa~outperforms the baselines in all the metrics. 
% , and model scaling brings significant performance gains.
% Note that no post-processing, e.g., CLIP reranking~\cite{dalle}, image caption reranking~\cite{cogview}, etc., is applied to our model. 
% reranking 大家都做了 但是cogview额外调过生成图片的对比度
Note that increasing the sampling size during inference is expected to bring clear improvements on FID and IS.
Compared with DALLE~\cite{dalle}, CogView~\cite{cogview} and N\"UWA~\cite{nvwa}, whose sampling sizes are $512$, $60$ and $60$, respectively, \ofa~outperforms these SOTA methods on FID and IS with a much smaller sampling size $24$. 
% In particular,  while DALLE~\cite{dalle} has shown clear improvements in FID and IS with the increase in sample size, our model outperforms previous SOTA methods with a smaller one.
This illustrates that \ofa~has learned better correspondence among the query text, the image and the image codes.

\begin{table}[t]
\center
\caption{Experimental results on text-to-image generation. Models are evaluated on FID, CLIPSIM, and IS scores. \ofa ~outperforms the baselines, including the concurrent SOTA N\"UWA. We report the results of $\text{OFA}_{\rm Large}$. 
% ``\#Sample'' refers to the sampling size during inference. 
Note that GLIDE additionally has $1.5B$ parameters for upsampling except for the $3.5B$ parameters. 
% ``\#Sample'' refers to the number of generated samples for post-selection. 
}
\vskip 0.15in
\begin{adjustbox}{max width=1.\textwidth}
\begin{tabular}{@{}lccccc@{}}
\toprule
  Model
%   & \#Sample 
  & FID$\downarrow$  & CLIPSIM$\uparrow$ & IS$\uparrow$
  \\
\midrule
  DALLE 
  \cite{dalle}
%   & 250M
%   & 
%   & 512
  & 27.5
  & -
  & 17.9

  \\
  CogView 
  \cite{cogview}
%   & 30M
%   &
%   & 60
  & 27.1 
  & 33.3
%   &0.3325
  &18.2
      \\
  GLIDE 
  \cite{nichol2021glide}
%   & 250M
%   & 
%   +1.5B
%   & -
  & 12.2
  & -
    & -
    \\
  Unifying 
  \cite{huang2021unifying}
%   & -
%   & $\surd$
%   & -
  & 29.9
  & 30.9
%   &0.3088
  &-
  \\
  N\"UWA
  \cite{nvwa}
%   & 2.9M
%   & $\surd$
%   & 60
  & 12.9 
  & 34.3
%   &0.3429
  &27.2

%   &0.3429

  \\
\midrule
%   $\text{OFA}\rm_{Base}$
% %   & 14M
% %   & $\surd$
%   & 180M
% %   & 24
%   & \textbf{20.8} 
%   & \textbf{31.6}
% %   &0.338
%   & \textbf{21.8}
%   \\
  $\text{OFA}$
%   & 14M
%   & $\surd$
%   & 24
  & \textbf{10.5} 
  & \textbf{34.4}
%   &0.338
  & \textbf{31.1}
  \\
\bottomrule
\end{tabular}
\end{adjustbox}
\label{tb:image-gen-results}
\vskip -0.1in
\end{table}

% \begin{table*}[t]
% \begin{adjustbox}{}
% \begin{tabular}{@{}lccccccc@{}}
% \toprule
%   Model & FID-0$\downarrow$ & FID-1$\downarrow$ & FID-2$\downarrow$ & FID-4$\downarrow$ & FID-8$\downarrow$ & CLIPSi$\uparrow$ & IS$\uparrow$
%   \\
% \midrule
%   DALLE \cite{ramesh2021zero}
%   & 27.5
%   &28.0 &45.5 &83.5& 85.0
%   & -
%   & 17.9
%   \\
%   Unifying \cite{huang2021unifying}
%   &29.9 &- &- &-& -
%   & 30.9
% %   &0.3088
%   &-
%   \\
%   CogView \cite{ding2021cogview}
%   &27.1 & 19.4 & 13.9& 19.4& 23.6
%   & 33.3
% %   &0.3325
%   &18.2
%   \\
%   NvWA \cite{wu2021n}
%   &12.9 & 13.8& 15.7& 19.3& 24
%   & 34.3
% %   &0.3429
%   &27.2
%   \\
% \midrule
%   \ofa
%   &10.7 & 9.59 & 10.4 &8.5 & 4.18
%   & 34.4
% %   &0.338
%   &31.1
%   \\
% \bottomrule
% \end{tabular}
% \end{adjustbox}
% \caption{Image Generation results}
% \label{tb:image-gen-results}
% \end{table*}
We compare \ofa~with CogView
% ~\footnote{\url{https://wudao.aminer.cn/CogView/index.html}. 
% Note that this API samples 8 images of resolution of $512 \times 512$ for each query. We select the first one of generated images and resize it to the resolution of $256 \times 256$.}
and GLIDE
% ~\footnote{\url{https://colab.research.google.com/drive/1q6tJ58UKod1eCOkbaUNGzF3K5BbXlB5m}. Note that the only publicly available GLIDE model is of base size ($\sim$385M).} 
on generation quality with normal and counterfactual queries.\footnote{For more implementation details, please refer to Appendix~\ref{sec:appendix_downstream}}
Normal queries describe existing things in the real world, while counterfactual queries refer to those describing things that could only exist in our imagination. 
For normal queries, both CogView and \ofa~generate images semantically consistent with the given texts, in comparison with GLIDE. The generated examples from our model can provide more sophisticated details of objects, say the horse and the double-decker bus. 
For counterfactual queries, we find that \ofa~is the only one that can generate the three imaginary scenes, which indicates its imaginative power based on its strong capability to align text to the image. 
See Appendix~\ref{sec:qualitative_examples} for more qualitative examples.

\subsection{Results on Uni-modal Tasks}
As the design of OFA unifies different modalities, we evaluate its performance on unimodal tasks, namely tasks of natural language and computer vision. 
For natural language tasks, we evaluate OFA on $6$ tasks of the GLUE benchmark~\cite{glue} for natural language understanding and Gigaword abstractive summarization~\cite{gigaword} for natural language generation. 
For computer vision, we evaluate OFA on the classic ImageNet-1K~\cite{imagenet} dataset for image classification. 
More details are provided in Appendix~\ref{sec:appendix_downstream}.

\begin{table}[t]
\caption{Experimental results on the GLUE benchmark datasets~\cite{glue}. 
For comparison, we list the performance of multimodal pretrained models as well the recent SOTA models that were pretrained on natural language data only. Following \cite{roberta}, we finetune RTE and MRPC starting from the checkpoint finetuned on MNLI. 
% All models are of \textit{Base} size, and they share a similar amount of parameters. 
% For models of \textit{Base} size, \ofa~outperforms all multimodal pretrained baseline models on the $7$ tasks, and BERT on $6$ tasks except CoLA. For models of \textit{Large} size, \ofa~performs competitively with $\rm BERT_{Large}$. The reported results of the multimodal baselines are from \cite{iki2021effect} and the corresponding original papers. 
}
\vskip 0.15in
\centering
\begin{adjustbox}{max width=1.\textwidth}
\begin{tabular}{@{}lcccccc@{}}
\toprule
  Model
%   &CoLA
  &SST-2
  &RTE
  &MRPC
  &QQP
  &MNLI
  &QNLI
  \\
\midrule
\multicolumn{4}{l}{\textit{Multimodal Pretrained Baseline Models}}
%   &
  &
  &
  &
%   \\
%   BERT
%   &\textbf{54.6}
%   &92.5
%   &62.5
%   &81.9/87.6
%   &90.6/87.4
%   &84.2
%   &91.0
  \\
  VisualBERT~\cite{visualbert}
%   &38.6
  &89.4
  &56.6
  &71.9
  &89.4
  &81.6
  &87.0
  \\
  UNITER~\cite{uniter}
%   &37.4
  &89.7
  &55.6
  &69.3
  &89.2
  &80.9
  &86.0
  \\
  VL-BERT~\cite{vlbert}
%   &38.7
  &89.8
  &55.7
  &70.6
  &89.0
  &81.2
  &86.3
  \\
  VilBERT~\cite{vilbert}
%   &36.1
  &90.4
  &53.7
  &69.0
  &88.6
  &79.9
  &83.8
  \\
  LXMERT~\cite{lxmert}
%   &39.0
  &90.2
  &57.2
  &69.8
  &75.3
  &80.4
  &84.2
  \\
  Uni-Perceiver~\cite{uni-perceiver}
  &90.2
  &64.3
  &86.6
  &87.1
  &81.7
  &89.9
  \\
  SimVLM~\cite{simvlm}
%   &46.7
  &90.9
  &63.9
  &75.2
  &90.4
  &83.4
  &88.6
  \\
  FLAVA~\cite{flava}
%   &50.7
  &90.9
  &57.8
  &81.4
  &90.4
  &80.3
  &87.3
    \\
  UNIMO~\cite{unimo}
%   &39.0
  &96.8
  &-
  &-
  &-
  &89.8
  &-
  \\
  \midrule
\multicolumn{5}{l}{\textit{Natural-Language-Pretrained SOTA Models}}
%   &
  &
  &
  \\
  BERT~\cite{bert}
%   &60.6
  &93.2
  &70.4
  &88.0
  &91.3
  &86.6
  &92.3
  \\
  RoBERTa~\cite{roberta}
%   &68.0
  &96.4
  &86.6
  &90.9
  &92.2
  &90.2
  &93.9
  \\
  XLNet~\cite{xlnet}
%   &69.0
  &\textbf{97.0}
  &85.9
  &90.8
  &92.3
  &90.8
  &94.9
  \\
  ELECTRA~\cite{electra}
%   &69.1
  &96.9
  &88.0
  &90.8
  &92.4
  &90.9
  &95.0
  \\
  DeBERTa~\cite{deberta}
%   &70.5
  &96.8
  &88.3
  &\textbf{91.9}
  &92.3
  &\textbf{91.1}
  &\textbf{95.3}
  \\
%   UniLM
%   &\textbf{61.1}
%   &94.5
%   &70.9
%   &-/-
%   &-/-
%   &\textbf{87.0}
%   &92.7
%   \\
\midrule
\multicolumn{4}{l}{\textit{Ours}}
%   &
  &
  &
  &
  \\
  \ofa
%   &-
  & 96.6
  & \textbf{91.0}
  & 91.7
  & \textbf{92.5}
  & 90.2
  & 94.8
  \\
\bottomrule
\end{tabular}
\end{adjustbox}
\label{tb:glue-results}
\end{table}
\begin{table}[t]
\caption{Experimental results on Gigaword abstractive summarization. We report performance on the ROUGE evaluation~\cite{rouge}}.
\vskip 0.15in
\center
\begin{adjustbox}{max width=1.\textwidth}
\begin{tabular}{lcccc}
\toprule
  \multirow{2}*{Model}
  & \multicolumn{3}{c}{Gigaword}
  \\
  & ROUGE-1 & ROUGE-2 & ROUGE-L
  \\
\midrule
  BERTSHARE~\cite{bertshare}
  & 38.13 & 19.81 & 35.62
  \\
  MASS~\cite{mass}
  & 38.73 & 19.71 & 35.96
  \\
  UniLM~\cite{unilm}
  & 38.45 & 19.45 & 35.75
  \\
  PEGASUS~\cite{pegasus}
  & 39.12 & 19.86 & 36.24
  \\ 
  ProphetNet~\cite{prophetnet}
  & 39.55 & 20.27 & 36.57
  \\ 
  UNIMO~\cite{unimo}
  & 39.71 & 20.37 & 36.88
  \\ 
\midrule
  \ofa
  & \textbf{39.81} & \textbf{20.66} & \textbf{37.11}
  \\
\bottomrule
\end{tabular}
\end{adjustbox}
\label{tb:nlg-results}
\vskip -0.1in
\end{table}
As OFA has been pretrained on plain text data, it can be directly transferred to natural language downstream tasks. For natural language generation, it is essentially a sequence-to-sequence generation task, and for natural language understanding, typically text classification, we regard them as generation tasks where labels are essentially word sequences. 
% To be specific, we apply the all-candidate evaluation mentioned above for the classification tasks. 
Additionally, for each task, we design a manual instruction to indicate the model what types of questions it should answer. We list our instruction design in Appendix~\ref{sec:appendix_downstream}.

We demonstrate that even a unified multimodal pretrained model can achieve highly competitive performance in natural language tasks. Specifically, in the evaluation of natural language understanding, OFA surpasses multimodal pretrained models by large margins in all tasks. In comparison with the state-of-the-art natural language pretrained models, including RoBERTa~\cite{roberta}, XLNET~\cite{xlnet}, ELECTRA~\cite{electra}, and DeBERTa~\cite{deberta}, OFA reaches a comparable performance. In the evaluation of natural language generation, OFA even reaches a new state-of-the-art performance on the Gigaword dataset. 

Also, OFA can reach a competitive performance in image classification. Table~\ref{tb:image-classify-results} shows the performance of \ofa~on image classification. $\text{OFA}\rm_{Large}$~achieves higher accuracy than previous backbone models such as EfficientNet-B7~\cite{tan2019efficientnet} and ViT-L~\cite{vit}. We also compare \ofa~with self-supervised pretraining models based on contrastive learning and masked image modeling. \ofa~outperforms contrastive-based models such as SimCLR~\cite{simclr} and MoCo-v3~\cite{moco, mocov3} with similar parameters. Compared with pretrained models based on masked image modeling, e.g., BEiT-L~\cite{beit} and MAE-L~\cite{mae}, \ofa~can achieve similar performance. 

These aforementioned results in both natural language and vision tasks indicate that a unified multimodal pretrained model is not only effective in multimodal tasks but also capable of tackling unimodal tasks, and in the future, it might be sufficient for such a model to solve complex tasks concerning different modality combinations.

\begin{table}[t]
\center
\caption{ImageNet-1K finetuning results. All the listed models do not use extra labeled image classification samples during training for fair comparison. We report the results of $\text{OFA}\rm_{Large}$.} 
\vskip 0.15in
\begin{adjustbox}{max width=1.\textwidth}
\begin{tabular}{@{}lc@{}}
% 注：
% 1. EfficientNet-B7, DINO不存在和\ofa在comparable model-size的实验结果，这里列出的是小模型结果，考虑是否还保留
% 2. ViT这里的结果采用了MAE的reimplementation，原文的结果低得不太正常（76.2），MAE的结果更加可靠一些
% 3. concern：是否需要将baseline按照无pretrain、CL pretrain和MIM pretrain区分开？
% 4. 这里MAE结果低于BEiT，主要是我们用的是384分辨率的BEiT结果，MAE只有224分辨率的结果
\toprule
  Model & Top-1 Acc.
  \\
\midrule
  EfficientNet-B7 \cite{tan2019efficientnet}
  & 84.3
  \\    
  ViT-L/16 \cite{vit}
  & 82.5
  \\ 
  DINO \cite{dino}
  & 82.8
  \\    
  SimCLR v2 \cite{simclr}
  & 82.9
  \\  
  MoCo v3 \cite{mocov3}
  & 84.1
  \\   
  BEiT$_{384}$-L/16 \cite{beit}
  & \textbf{86.3}
  \\    
  MAE-L/16 \cite{mae}
  & 85.9 
  \\   
\midrule
  $\text{OFA}$
  & 85.6
  \\
\bottomrule
\end{tabular}
\end{adjustbox}
\label{tb:image-classify-results}
\vskip -0.1in
\end{table}

\subsection{Zero-shot Learning \& Task Transfer}
The instruction-guided pretraining enables \ofa~to perform zero-shot inference. Following Uni-Perceiver~\cite{uni-perceiver}, we evaluate our model on the $6$ tasks of the GLUE benchmark, including single-sentence classification and sentence pair classification.
Table~\ref{tb:zero-shot-glue-results} demonstrates that \ofa~generally outperforms Uni-Perceiver. 
However, both models do not achieve satisfactory performance in sentence-pair classification (with $\text{Acc}. < 60\%$). 
We hypothesize that the missing sentence-pair data in the pretraining dataset attributes to the performance.  

Also, we find that the model performance is highly sensitive to the design of instructions. To obtain the best result, one should search a proper instruction template possibly from a large pool of candidates. 
A slight change to manual prompts or model parameters may drastically influence the model performance, which is not robust. We leave this issue to the future work.

\begin{figure}[t]
    \centering
    \includegraphics[width=.9\linewidth]{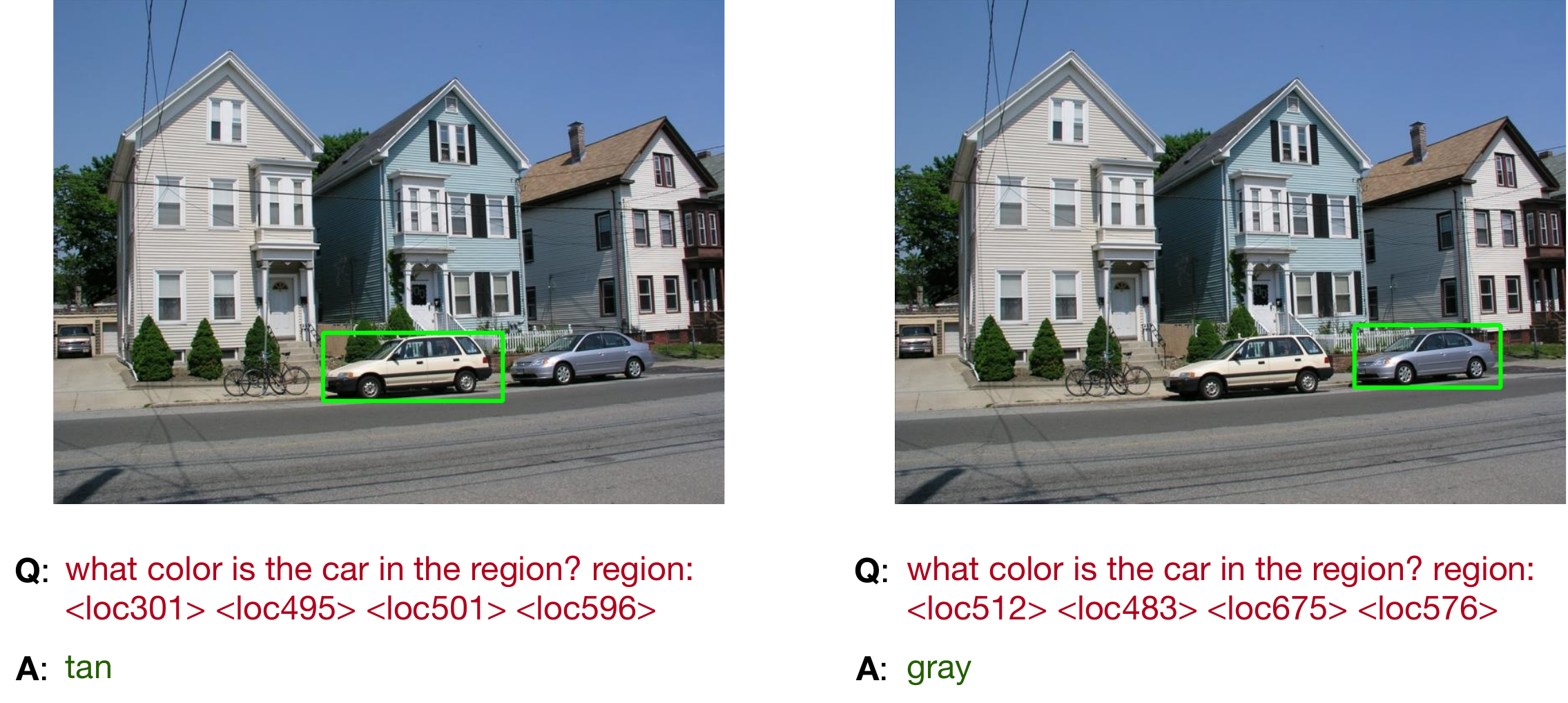}
    \caption{Qualitative results on an unseen task grounded QA. We design a new task called grounded question answering, where the model should answer a question about a certain region in the image. More samples are provided in Figure~\ref{fig:instruction_case_appendix} in Appendix~\ref{sec:qualitative_examples}.}
    \label{fig:grounded_qa_samples}
\end{figure}

\begin{figure}[t]
    \centering
    \includegraphics[width=.9\linewidth]{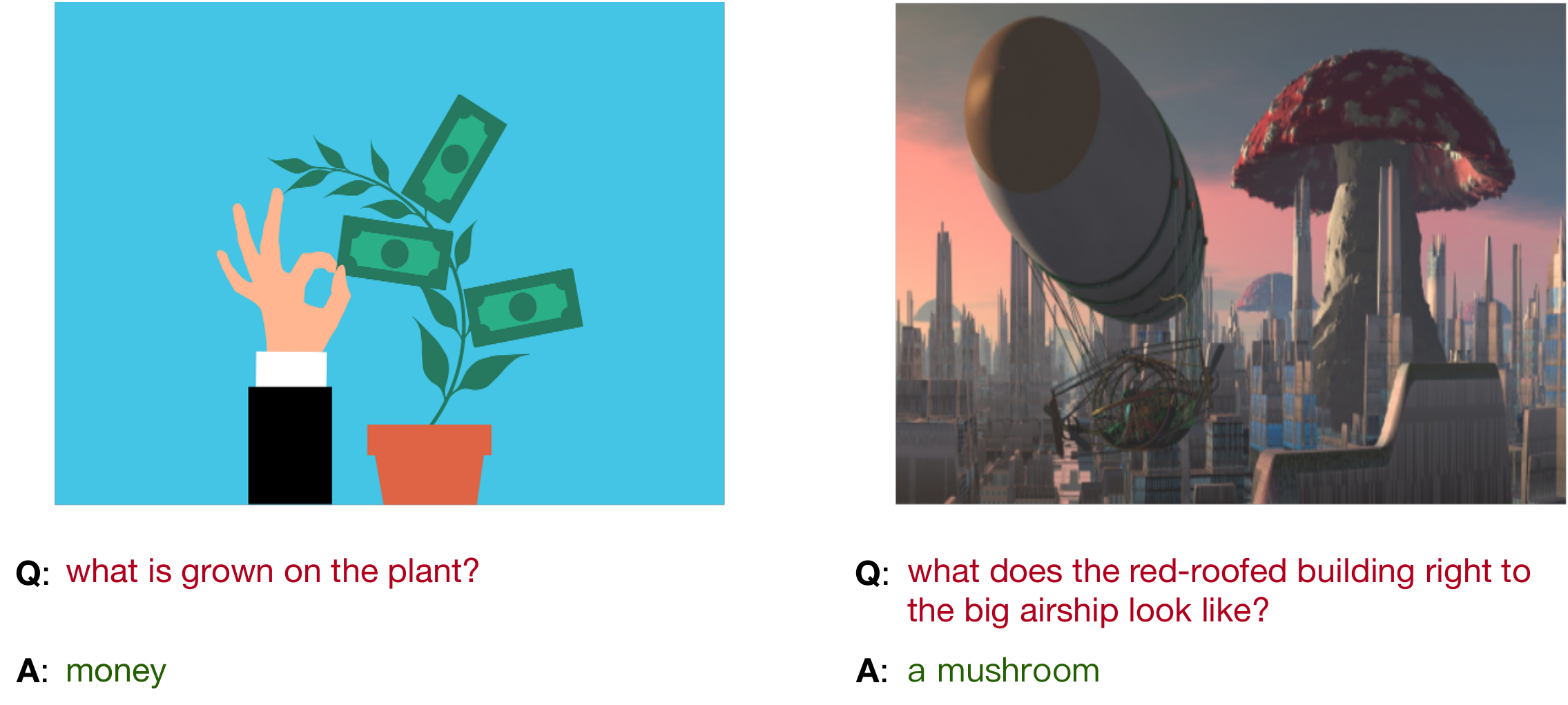}
    \caption{Qualitative results on unseen domain VQA. During pretraining, only real-world photographs are used for VQA. We present cases of VQA on out-of-domain images, i.e., the iconic and sci-fi images, and demonstrate their capability of transferring to unseen domains. More samples are provided in Figure~\ref{fig:ood_case_appendix} in Appendix~\ref{sec:qualitative_examples}.}
    \label{fig:ood_qa_samples}
\end{figure}

% \paragraph{Qualitative Examples}
\label{subsec:case_study}
% \zc{4.6 seems to be a different-level section. Too many negative words for competitors.}
% We compare \ofa~with CogView
% % ~\footnote{\url{https://wudao.aminer.cn/CogView/index.html}. 
% % Note that this API samples 8 images of resolution of $512 \times 512$ for each query. We select the first one of generated images and resize it to the resolution of $256 \times 256$.}
% and GLIDE
% % ~\footnote{\url{https://colab.research.google.com/drive/1q6tJ58UKod1eCOkbaUNGzF3K5BbXlB5m}. Note that the only publicly available GLIDE model is of base size ($\sim$385M).} 
% on the generation quality with normal queries and counterfactual queries.\footnote{For more implementation details, please refer to Appendix~\ref{sec:appendix_downstream}}
% Normal queries refer to those describing existing things in the real world, while counterfactual queries refer to those describing things that could only exist in our imagination. 
% For normal queries, both CogView and \ofa~generate images semantically consistent with the given texts, in comparison with GLIDE. The generated examples from our model can provide more sophisticated details of objects, say the horse and the double-docker bus. 
% For counterfactual queries, we find that \ofa~is the only one that can generate the three imaginary scenes, which indicates its imaginative power based on its strong capability to align text to image. 
% We strongly suggest readers referring to Appendix~\ref{sec:qualitative_examples} for more qualitative examples.
% TODO vqa cases
We observe that the model can transfer to unseen tasks well with new task instructions. We design a new task called grounded question answering and present examples in Figure~\ref{fig:grounded_qa_samples}. In this scenario, given a question about a certain region on the image, the model should provide a correct answer. We find that the model can achieve a satisfactory performance in this new task, which reflects its strong transferability. 
Besides, \ofa~can solve tasks with the out-of-domain input data. 
For example, \ofa~without finetuning achieves satisfactory performance in VQA for the out-of-domain images. Examples are demonstrated in Figure~\ref{fig:ood_qa_samples}.
\ofa~can also perform accurate visual grounding on the out-of-domain images, e.g., anime pictures, synthetic images, etc., and we demonstrate more examples on Figure~\ref{fig:vg_samples} in Appendix~\ref{sec:qualitative_examples}. 
% For text-to-image synthesis, our cases in Figure~\ref{fig:image_gen_samples} shows that \ofa~generates high-quality samples with counterfactual queries. We demonstrate more examples in Figure~\ref{fig:image_gen_samples2} in Appendix~\ref{sec:qualitative_examples}. 
% We compare the images generated by \ofa and Cogview in Figure \ref{fig:image_gen_samples}. 
% Given the normal queries, although the images generated by CogView are match these queries, they are much worse than the images generated by \ofa in terms of authenticity and clear.
% In addition, Cogview does not perform well given the counterfactual queries. In contrast, the images generated by \ofa are very consistent with these queries, indicating the superior cross-modal understanding and generation abilities of \ofa.
\begin{table}[t]
\caption{Zero-shot performance on $6$ GLUE tasks and SNLI-VE.}
\vskip 0.15in
\center
\begin{adjustbox}{max width=1.\textwidth}
\begin{tabular}{@{\extracolsep{\fill}}lccccccc}
\toprule
  \multirow{2}*{Model}
  &SST-2
  &RTE
  &MRPC
  &QQP
  &QNLI
  &MNLI
  &SNLI-VE
  \\
  &{Acc.}
  &{Acc.}
  &{F1}
  &{F1}
  &{Acc.}
  &{Acc.}
  &{Acc. (dev/test)}
  \\  
\midrule
  Uni-Perceiver
%   ~\cite{uni-perceiver}
  &70.6
  &55.6
  &76.1
  &53.6
  &51.0
  &\textbf{49.6}
  &-
  \\
\midrule
  $\text{\ofa}_{\rm Base}$
  &\textbf{71.6}
  &\textbf{56.7}
  &\textbf{79.5}
  &\textbf{54.0}
  &\textbf{51.4}
  &37.3
  &\textbf{49.71 / 49.18}
  \\
\bottomrule
\end{tabular}
\end{adjustbox}
\label{tb:zero-shot-glue-results}
\end{table}

\subsection{Ablation on Multitask Pretraining}
Thanks to the unified framework, \ofa~has been pretrained on multiple tasks and thus endowed with comprehensive capabilities. 
However, the effects of each task are still undiscovered. 
We verify their effects on multiple downstream tasks, including image captioning, VQA, image classification, and text-to-image generation.
% Previous studies have provided evidence about the contribution of conventional pretraining tasks, e.g., MLM, MOC, ITM, VQA, etc.
%  the evaluation of uni-modal pretraining tasks on multimodal pretraining, and also detection and grounded captioning. 

% To fill in this blank, we verify the key pretraining tasks in our setup by conducting an ablation study to evaluate how each task influences the downstream performance of \ofa. Specifically, 
% Additionally, we also evaluate the impacts from Trie-tree search. 

% We validate the effectiveness of model components by conducting ablation experiments. 
% The ablation covers the design of pretraining tasks. An $\rm \ofa_{Base}$ model with its ablated variants are evaluated on image captioning, VQA, image classification and text-to-image generation, which provides comprehensive evidence to various types of downstream tasks.

%In this section, we validate the effectiveness of each model component by conducting ablation study on image captioning, visual question answersing, image classification and image generation tasks. To save experimental cost, all models are of Base size. 
%First, we explore the effectiveness of text-only data. We find that using text-only data is beneficial for fine-grained multimodal tasks such as image captioning and vision question answering, but it harmful to image classification task. This suggest that xxx.
% 这块不知道咋说，标记一下先。。

We first evaluate how uni-modal pretraining tasks influence the performance in both cross-modal and uni-modal tasks. 
Table \ref{tb:ablation-results} demonstrates our experimental results. We observe some interesting phenomena about the effects of uni-modal pretraining tasks. 
Text infilling brings improvement on image caption ($+0.8$ CIDEr) and VQA ($+0.46$ Acc.). 
Natural language pretraining betters the contextualized representation of language and thus enhances performance in cross-modal tasks. 
However, it is noticed that the language pretraining task may degrade the performance in image classification, leading to the decrease in ImageNet-1K ($-1.0$ Acc.). 
Also, it is interesting to find that it does not encourage improvement in text-to-image generation ($-0.1$ CLIPSIM). 
It may attribute to the simplicity of text in this task, which indicates that improved representation of language does not affect the performance. 
As to image infilling, it significantly improves the performance in image classification ($+1.0$ Acc.) and text-to-image generation ($+0.6$ CLIPSIM). Learning to recover images is an effective self-supervised task for image representation, and it also encourages the decoder's ability to generate image codes. 
However, it hurts the performance in image captioning and VQA. Both tasks require a strong capability in generating texts, and the decoder's learning of image generation naturally brings performance degradation in captioning ($-0.7$ CIDEr) and VQA ($-0.3$ Acc.).

\begin{table}[t]
\caption{Ablation results of \ofa. All models are pretrained for $250k$ steps. \textit{w/o ground.} represents the removal of both visual grounding and grounded captioning tasks. Note that all models are only finetuned with the cross-entropy loss in image captioning.}
\vskip 0.15in
\centering
\begin{adjustbox}{max width=1.\textwidth}
\begin{tabular}{@{}lcccccc@{}}
\toprule
  \multirow{2}*{Model}
  &Caption
  &VQA
  &ImageNet
  &Image Generation
  \\
  &{CIDEr}
  &{Test-dev}
  &{Top-1 Acc.}
  &{FID / CLIPSIM / IS}
  \\
\midrule
  $\text{OFA}\rm_{Base}$
  &135.6
  &76.0
  &82.2
  &20.8 / 31.6 / 21.5
  \\
  \midrule
%   \ \ \textit{w/o text-only data}
  \ \ \textit{w/o text infill.}
  &134.8
  &75.6
  &83.2
  &20.3 / 31.7 / 21.8
  \\
%   \ \ \textit{w/o image-only data}
\ \ \textit{w/o image infill.}
  &136.3
  &76.3
  &81.8
  &23.2 / 31.0 / 20.0
  \\
%   \ \ \textit{w/o detection \& grounding}
  \ \ \textit{w/o det.}
  &133.3
  &75.4
  &81.4
  & 20.9 / 31.5 / 21.6
  \\
  \ \ \textit{w/o ground.}
  &134.2
  &75.5
  &82.0
  & 21.2 / 31.5 / 21.5
  \\
%   \ \ \textit{w/o downstream.}
%   &132.1
%   &74.02
%   &81.2
%   & 20.9  / 31.5 / 21.7
%   \\
%   \ \ \textit{w/o Normformer}
%   &133.8
%   &
%   &81.3
%   &
%   \\
%   \ \ \textit{w/o Trie-based search}
% \midrule
%   \ \ \textit{w/o Trie}
%   &-
%   &75.86
%   &81.9
%   &-
%   \\
\bottomrule
\end{tabular}
\end{adjustbox}
\label{tb:ablation-results}
\end{table}

Furthermore, we evaluate how multimodal tasks impact the performance. 
Previous studies have provided evidence of the contribution of conventional pretraining tasks, e.g., MLM, MOC, ITM, VQA, image captioning, etc.~\cite{uniter, vinvl}. 
However, they miss other tasks, e.g., detection and visual grounding \& grounded captioning. 
We conduct experiments on these tasks and find that tasks predicting regions are crucial to multimodal tasks, with a performance increase in image captioning ($+2.3$ CIDEr \& $+1.4$ CIDEr) and VQA ($+0.6$ Acc. \& $+0.5$ Acc.). 
It suggests that detection and visual grounding \& grounded captioning help the model grasp fined-grained alignments between vision and language. 
Region information contributes little to text-to-image generation ($+0.1$ CLIPSIM \& $+0.1$ CLIPSIM), as this task requires far less text-region alignment information. 
We surprisingly find that detection can encourage the performance in visual understanding ($+0.8$ Acc.). It indicates that incorporating region information might be essential to visual understanding, especially on images with complex objects.

\section{Conclusion}
In this work, we propose \textbf{\ofa}, a Task-Agnostic and Modality-Agnostic framework supporting Task Comprehensiveness. 
\ofa~achieves the unification in architecture, tasks and modalities, and thus is capable of multimodal \& uni-modal understanding and generation, without specification in additional layers or tasks. 
Our experiments show that \ofa~creates new SOTAs in a series of tasks, including image captioning, VQA, visual entailment, and referring expression comprehension. \ofa~also demonstrates a comparable performance with language / vision pretrained SOTA models in uni-modal understanding and generation tasks, e.g., GLUE, abstractive summarization, and image classification. 
We provide a further analysis to demonstrate its capability in zero-shot learning and domain \& task transfer, and we also verify the effectiveness of pretraining tasks.  

In the future, we will continue exploring the issues discovered in this work. Also, we endeavor to figure out a reasonable solution to building an omni-model essentially generalizable to the complex real world. 

% the effects of proposed pretraining tasks as well as strong capability in zeroshot learning. In our case study, we also provide a series of qualitative examples to verify the effectiveness of \ofa~in text-to-image generation, visual grounding, and open-domain VQA. 
\label{sec:conclusion}

\section*{Acknowledgments}
We would like to thank Jie Zhang, Yong Li, Jiamang Wang, Shao Yuan, and Zheng Cao for their support to this project, and we would like to thank Guangxiang Zhao and Fei Sun for their insightful comments to our paper. 

%Bibliography
\bibliographystyle{unsrt}  
\bibliography{main}  

\begin{thebibliography}{100}

\bibitem{transformer}
Ashish Vaswani, Noam Shazeer, Niki Parmar, Jakob Uszkoreit, Llion Jones,
  Aidan~N. Gomez, Lukasz Kaiser, and Illia Polosukhin.
\newblock Attention is all you need.
\newblock In {\em NeurIPS 2017}, pages 5998--6008, 2017.

\bibitem{bert}
Jacob Devlin, Ming{-}Wei Chang, Kenton Lee, and Kristina Toutanova.
\newblock {BERT:} pre-training of deep bidirectional transformers for language
  understanding.
\newblock In Jill Burstein, Christy Doran, and Thamar Solorio, editors, {\em
  {NAACL-HLT} 2019}, pages 4171--4186. Association for Computational
  Linguistics, 2019.

\bibitem{gpt3}
Tom~B Brown, Benjamin Mann, Nick Ryder, Melanie Subbiah, Jared Kaplan, Prafulla
  Dhariwal, Arvind Neelakantan, Pranav Shyam, Girish Sastry, Amanda Askell,
  et~al.
\newblock Language models are few-shot learners.
\newblock {\em arXiv preprint arXiv:2005.14165}, 2020.

\bibitem{transformer_math}
Karl Cobbe, Vineet Kosaraju, Mohammad Bavarian, Jacob Hilton, Reiichiro Nakano,
  Christopher Hesse, and John Schulman.
\newblock Training verifiers to solve math word problems.
\newblock {\em arXiv preprint arXiv:2110.14168}, 2021.

\bibitem{wav2vec}
Steffen Schneider, Alexei Baevski, Ronan Collobert, and Michael Auli.
\newblock wav2vec: Unsupervised pre-training for speech recognition.
\newblock {\em arXiv preprint arXiv:1904.05862}, 2019.

\bibitem{vit}
Alexey Dosovitskiy, Lucas Beyer, Alexander Kolesnikov, Dirk Weissenborn,
  Xiaohua Zhai, Thomas Unterthiner, Mostafa Dehghani, Matthias Minderer, Georg
  Heigold, Sylvain Gelly, et~al.
\newblock An image is worth 16x16 words: Transformers for image recognition at
  scale.
\newblock {\em arXiv preprint arXiv:2010.11929}, 2020.

\bibitem{perceiver}
Andrew Jaegle, Felix Gimeno, Andrew Brock, Andrew Zisserman, Oriol Vinyals, and
  Joao Carreira.
\newblock Perceiver: General perception with iterative attention.
\newblock {\em arXiv preprint arXiv:2103.03206}, 2021.

\bibitem{vlbert}
Weijie Su, Xizhou Zhu, Yue Cao, Bin Li, Lewei Lu, Furu Wei, and Jifeng Dai.
\newblock Vl-bert: Pre-training of generic visual-linguistic representations.
\newblock In {\em International Conference on Learning Representations}, 2019.

\bibitem{flan}
Jason Wei, Maarten Bosma, Vincent~Y Zhao, Kelvin Guu, Adams~Wei Yu, Brian
  Lester, Nan Du, Andrew~M Dai, and Quoc~V Le.
\newblock Finetuned language models are zero-shot learners.
\newblock {\em arXiv preprint arXiv:2109.01652}, 2021.

\bibitem{t0}
Victor Sanh, Albert Webson, Colin Raffel, Stephen~H Bach, Lintang Sutawika,
  Zaid Alyafeai, Antoine Chaffin, Arnaud Stiegler, Teven~Le Scao, Arun Raja,
  et~al.
\newblock Multitask prompted training enables zero-shot task generalization.
\newblock {\em arXiv preprint arXiv:2110.08207}, 2021.

\bibitem{adapter}
Neil Houlsby, Andrei Giurgiu, Stanislaw Jastrzebski, Bruna Morrone, Quentin
  De~Laroussilhe, Andrea Gesmundo, Mona Attariyan, and Sylvain Gelly.
\newblock Parameter-efficient transfer learning for nlp.
\newblock In {\em International Conference on Machine Learning}, pages
  2790--2799. PMLR, 2019.

\bibitem{prompt_tuning}
Brian Lester, Rami Al-Rfou, and Noah Constant.
\newblock The power of scale for parameter-efficient prompt tuning.
\newblock {\em arXiv preprint arXiv:2104.08691}, 2021.

\bibitem{vilbert}
Jiasen Lu, Dhruv Batra, Devi Parikh, and Stefan Lee.
\newblock Vilbert: Pretraining task-agnostic visiolinguistic representations
  for vision-and-language tasks.
\newblock In {\em NeurIPS}, 2019.

\bibitem{uniter}
Yen-Chun Chen, Linjie Li, Licheng Yu, Ahmed~El Kholy, Faisal Ahmed, Zhe Gan,
  Yu~Cheng, and Jingjing Liu.
\newblock Uniter: Universal image-text representation learning.
\newblock In {\em ECCV}, 2020.

\bibitem{oscar}
Xiujun Li, Xi~Yin, Chunyuan Li, Xiaowei Hu, Pengchuan Zhang, Lei Zhang, Lijuan
  Wang, Houdong Hu, Li~Dong, Furu Wei, Yejin Choi, and Jianfeng Gao.
\newblock Oscar: Object-semantics aligned pre-training for vision-language
  tasks.
\newblock In {\em ECCV}, 2020.

\bibitem{villa}
Zhe Gan, Yen-Chun Chen, Linjie Li, Chen Zhu, Yu~Cheng, and Jingjing Liu.
\newblock Large-scale adversarial training for vision-and-language
  representation learning.
\newblock {\em ArXiv}, abs/2006.06195, 2020.

\bibitem{vinvl}
Pengchuan Zhang, Xiujun Li, Xiaowei Hu, Jianwei Yang, Lei Zhang, Lijuan Wang,
  Yejin Choi, and Jianfeng Gao.
\newblock Vinvl: Revisiting visual representations in vision-language models.
\newblock {\em 2021 IEEE/CVF Conference on Computer Vision and Pattern
  Recognition (CVPR)}, pages 5575--5584, 2021.

\bibitem{m6}
Junyang Lin, Rui Men, An~Yang, Chang Zhou, Ming Ding, Yichang Zhang, Peng Wang,
  Ang Wang, Le~Jiang, Xianyan Jia, et~al.
\newblock M6: A chinese multimodal pretrainer.
\newblock {\em arXiv preprint arXiv:2103.00823}, 2021.

\bibitem{ufc}
Zhu Zhang, Jianxin Ma, Chang Zhou, Rui Men, Zhikang Li, Ming Ding, Jie Tang,
  Jingren Zhou, and Hongxia Yang.
\newblock M6-ufc: Unifying multi-modal controls for conditional image
  synthesis.
\newblock {\em arXiv preprint arXiv:2105.14211}, 2021.

\bibitem{m6-t}
An~Yang, Junyang Lin, Rui Men, Chang Zhou, Le~Jiang, Xianyan Jia, Ang Wang, Jie
  Zhang, Jiamang Wang, Yong Li, et~al.
\newblock Exploring sparse expert models and beyond.
\newblock {\em arXiv preprint arXiv:2105.15082}, 2021.

\bibitem{m6-10t}
Junyang Lin, An~Yang, Jinze Bai, Chang Zhou, Le~Jiang, Xianyan Jia, Ang Wang,
  Jie Zhang, Yong Li, Wei Lin, et~al.
\newblock M6-10t: A sharing-delinking paradigm for efficient multi-trillion
  parameter pretraining.
\newblock {\em arXiv preprint arXiv:2110.03888}, 2021.

\bibitem{simvlm}
Zirui Wang, Jiahui Yu, Adams~Wei Yu, Zihang Dai, Yulia Tsvetkov, and Yuan Cao.
\newblock Simvlm: Simple visual language model pretraining with weak
  supervision.
\newblock {\em ArXiv}, abs/2108.10904, 2021.

\bibitem{florence}
Lu~Yuan, Dongdong Chen, Yi-Ling Chen, Noel C.~F. Codella, Xiyang Dai, Jianfeng
  Gao, Houdong Hu, Xuedong Huang, Boxin Li, Chunyuan Li, Ce~Liu, Mengchen Liu,
  Zicheng Liu, Yumao Lu, Yu~Shi, Lijuan Wang, Jianfeng Wang, Bin Xiao, Zhen
  Xiao, Jianwei Yang, Michael Zeng, Luowei Zhou, and Pengchuan Zhang.
\newblock Florence: A new foundation model for computer vision.
\newblock {\em ArXiv}, abs/2111.11432, 2021.

\bibitem{gpt}
Alec Radford, Karthik Narasimhan, Tim Salimans, and Ilya Sutskever.
\newblock Improving language understanding by generative pre-training.
\newblock {\em URL
  https://s3-us-west-2.amazonaws.com/openai-assets/researchcovers/
  languageunsupervised/language understanding paper. pdf}, 2018.

\bibitem{xlnet}
Zhilin Yang, Zihang Dai, Yiming Yang, Jaime~G. Carbonell, Ruslan Salakhutdinov,
  and Quoc~V. Le.
\newblock Xlnet: Generalized autoregressive pretraining for language
  understanding.
\newblock In {\em NeurIPS 2019}, pages 5754--5764, 2019.

\bibitem{ernie}
Yu~Sun, Shuohuan Wang, Yu{-}Kun Li, Shikun Feng, Xuyi Chen, Han Zhang, Xin
  Tian, Danxiang Zhu, Hao Tian, and Hua Wu.
\newblock {ERNIE:} enhanced representation through knowledge integration.
\newblock {\em CoRR}, abs/1904.09223, 2019.

\bibitem{ernie2.0}
Yu~Sun, Shuohuan Wang, Yu{-}Kun Li, Shikun Feng, Hao Tian, Hua Wu, and Haifeng
  Wang.
\newblock {ERNIE} 2.0: {A} continual pre-training framework for language
  understanding.
\newblock {\em CoRR}, abs/1907.12412, 2019.

\bibitem{roberta}
Yinhan Liu, Myle Ott, Naman Goyal, Jingfei Du, Mandar Joshi, Danqi Chen, Omer
  Levy, Mike Lewis, Luke Zettlemoyer, and Veselin Stoyanov.
\newblock Roberta: {A} robustly optimized {BERT} pretraining approach.
\newblock {\em CoRR}, abs/1907.11692, 2019.

\bibitem{unilm}
Li~Dong, Nan Yang, Wenhui Wang, Furu Wei, Xiaodong Liu, Yu~Wang, Jianfeng Gao,
  Ming Zhou, and Hsiao{-}Wuen Hon.
\newblock Unified language model pre-training for natural language
  understanding and generation.
\newblock In {\em NeurIPS 2019}, pages 13042--13054, 2019.

\bibitem{t5}
Colin Raffel, Noam Shazeer, Adam Roberts, Katherine Lee, Sharan Narang, Michael
  Matena, Yanqi Zhou, Wei Li, and Peter~J Liu.
\newblock Exploring the limits of transfer learning with a unified text-to-text
  transformer.
\newblock {\em Journal of Machine Learning Research}, 21(140):1--67, 2020.

\bibitem{bart}
Mike Lewis, Yinhan Liu, Naman Goyal, Marjan Ghazvininejad, Abdelrahman Mohamed,
  Omer Levy, Veselin Stoyanov, and Luke Zettlemoyer.
\newblock {BART}: Denoising sequence-to-sequence pre-training for natural
  language generation, translation, and comprehension.
\newblock In {\em ACL 2020}, July 2020.

\bibitem{simclr}
Ting Chen, Simon Kornblith, Mohammad Norouzi, and Geoffrey Hinton.
\newblock A simple framework for contrastive learning of visual
  representations.
\newblock In {\em International conference on machine learning}, pages
  1597--1607. PMLR, 2020.

\bibitem{moco}
Xinlei Chen, Haoqi Fan, Ross Girshick, and Kaiming He.
\newblock Improved baselines with momentum contrastive learning.
\newblock {\em arXiv preprint arXiv:2003.04297}, 2020.

\bibitem{byol}
Jean-Bastien Grill, Florian Strub, Florent Altch{\'e}, Corentin Tallec,
  Pierre~H Richemond, Elena Buchatskaya, Carl Doersch, Bernardo~Avila Pires,
  Zhaohan~Daniel Guo, Mohammad~Gheshlaghi Azar, et~al.
\newblock Bootstrap your own latent: A new approach to self-supervised
  learning.
\newblock {\em arXiv preprint arXiv:2006.07733}, 2020.

\bibitem{mocov3}
Xinlei Chen and Kaiming He.
\newblock Exploring simple siamese representation learning.
\newblock In {\em Proceedings of the IEEE/CVF Conference on Computer Vision and
  Pattern Recognition}, pages 15750--15758, 2021.

\bibitem{beit}
Hangbo Bao, Li~Dong, and Furu Wei.
\newblock Beit: Bert pre-training of image transformers.
\newblock {\em arXiv preprint arXiv:2106.08254}, 2021.

\bibitem{mae}
Kaiming He, Xinlei Chen, Saining Xie, Yanghao Li, Piotr Doll{\'a}r, and Ross
  Girshick.
\newblock Masked autoencoders are scalable vision learners.
\newblock {\em arXiv preprint arXiv:2111.06377}, 2021.

\bibitem{visualbert}
Liunian~Harold Li, Mark Yatskar, Da~Yin, Cho-Jui Hsieh, and Kai-Wei Chang.
\newblock Visualbert: A simple and performant baseline for vision and language.
\newblock {\em ArXiv}, abs/1908.03557, 2019.

\bibitem{vlp}
Luowei Zhou, Hamid Palangi, Lei Zhang, Houdong Hu, Jason~J. Corso, and Jianfeng
  Gao.
\newblock Unified vision-language pre-training for image captioning and {VQA}.
\newblock In {\em {AAAI} 2020}, pages 13041--13049, 2020.

\bibitem{lxmert}
Hao Tan and Mohit Bansal.
\newblock Lxmert: Learning cross-modality encoder representations from
  transformers.
\newblock In {\em Proceedings of the 2019 Conference on Empirical Methods in
  Natural Language Processing and the 9th International Joint Conference on
  Natural Language Processing (EMNLP-IJCNLP)}, pages 5100--5111, 2019.

\bibitem{unicoder-vl}
Gen Li, Nan Duan, Yuejian Fang, Daxin Jiang, and Ming Zhou.
\newblock Unicoder-vl: {A} universal encoder for vision and language by
  cross-modal pre-training.
\newblock {\em CoRR}, abs/1908.06066, 2019.

\bibitem{interbert}
Junyang Lin, An~Yang, Yichang Zhang, Jie Liu, Jingren Zhou, and Hongxia Yang.
\newblock Interbert: Vision-and-language interaction for multi-modal
  pretraining.
\newblock {\em arXiv preprint arXiv:2003.13198}, 2020.

\bibitem{vilbert-mt}
Jiasen Lu, Vedanuj Goswami, Marcus Rohrbach, Devi Parikh, and Stefan Lee.
\newblock 12-in-1: Multi-task vision and language representation learning.
\newblock In {\em Proceedings of the IEEE/CVF Conference on Computer Vision and
  Pattern Recognition}, pages 10437--10446, 2020.

\bibitem{e2e-vlp}
Haiyang Xu, Ming Yan, Chenliang Li, Bin Bi, Songfang Huang, Wenming Xiao, and
  Fei Huang.
\newblock E2e-vlp: End-to-end vision-language pre-training enhanced by visual
  learning.
\newblock {\em arXiv preprint arXiv:2106.01804}, 2021.

\bibitem{ernie_vil}
Fei Yu, Jiji Tang, Weichong Yin, Yu~Sun, Hao Tian, Hua Wu, and Haifeng Wang.
\newblock Ernie-vil: Knowledge enhanced vision-language representations through
  scene graphs.
\newblock In {\em Proceedings of the AAAI Conference on Artificial
  Intelligence}, volume~35, pages 3208--3216, 2021.

\bibitem{unimo}
Wei Li, Can Gao, Guocheng Niu, Xinyan Xiao, Hao Liu, Jiachen Liu, Hua Wu, and
  Haifeng Wang.
\newblock {UNIMO:} towards unified-modal understanding and generation via
  cross-modal contrastive learning.
\newblock In Chengqing Zong, Fei Xia, Wenjie Li, and Roberto Navigli, editors,
  {\em {ACL/IJCNLP} 2021}, pages 2592--2607. Association for Computational
  Linguistics, 2021.

\bibitem{pixelbert}
Zhicheng Huang, Zhaoyang Zeng, Bei Liu, Dongmei Fu, and Jianlong Fu.
\newblock Pixel-bert: Aligning image pixels with text by deep multi-modal
  transformers.
\newblock {\em ArXiv}, abs/2004.00849, 2020.

\bibitem{vlmo}
Wenhui Wang, Hangbo Bao, Li~Dong, and Furu Wei.
\newblock Vlmo: Unified vision-language pre-training with
  mixture-of-modality-experts.
\newblock {\em ArXiv}, abs/2111.02358, 2021.

\bibitem{clip}
Alec Radford, Jong~Wook Kim, Chris Hallacy, Aditya Ramesh, Gabriel Goh,
  Sandhini Agarwal, Girish Sastry, Amanda Askell, Pamela Mishkin, Jack Clark,
  Gretchen Krueger, and Ilya Sutskever.
\newblock Learning transferable visual models from natural language
  supervision.
\newblock In Marina Meila and Tong Zhang, editors, {\em {ICML} 2021}, volume
  139 of {\em Proceedings of Machine Learning Research}, pages 8748--8763.
  {PMLR}, 2021.

\bibitem{dalle}
Aditya Ramesh, Mikhail Pavlov, Gabriel Goh, Scott Gray, Chelsea Voss, Alec
  Radford, Mark Chen, and Ilya Sutskever.
\newblock Zero-shot text-to-image generation.
\newblock {\em arXiv preprint arXiv:2102.12092}, 2021.

\bibitem{cogview}
Ming Ding, Zhuoyi Yang, Wenyi Hong, Wendi Zheng, Chang Zhou, Da~Yin, Junyang
  Lin, Xu~Zou, Zhou Shao, Hongxia Yang, et~al.
\newblock Cogview: Mastering text-to-image generation via transformers.
\newblock {\em arXiv preprint arXiv:2105.13290}, 2021.

\bibitem{nvwa}
Chenfei Wu, Jian Liang, Lei Ji, Fan Yang, Yuejian Fang, Daxin Jiang, and Nan
  Duan.
\newblock N$\backslash$" uwa: Visual synthesis pre-training for neural visual
  world creation.
\newblock {\em arXiv preprint arXiv:2111.12417}, 2021.

\bibitem{vqvae}
A{\"a}ron van~den Oord, Oriol Vinyals, and Koray Kavukcuoglu.
\newblock Neural discrete representation learning.
\newblock In {\em NIPS}, 2017.

\bibitem{vqgan}
Patrick Esser, Robin Rombach, and Bjorn Ommer.
\newblock Taming transformers for high-resolution image synthesis.
\newblock In {\em Proceedings of the IEEE/CVF Conference on Computer Vision and
  Pattern Recognition}, pages 12873--12883, 2021.

\bibitem{kaiser2017one}
Lukasz Kaiser, Aidan~N Gomez, Noam Shazeer, Ashish Vaswani, Niki Parmar, Llion
  Jones, and Jakob Uszkoreit.
\newblock One model to learn them all.
\newblock {\em arXiv preprint arXiv:1706.05137}, 2017.

\bibitem{vlt5}
Jaemin Cho, Jie Lei, Haochen Tan, and Mohit Bansal.
\newblock Unifying vision-and-language tasks via text generation.
\newblock In {\em ICML}, 2021.

\bibitem{unicorn}
Zhengyuan Yang, Zhe Gan, Jianfeng Wang, Xiaowei Hu, Faisal Ahmed, Zicheng Liu,
  Yumao Lu, and Lijuan Wang.
\newblock Crossing the format boundary of text and boxes: Towards unified
  vision-language modeling.
\newblock {\em ArXiv}, abs/2111.12085, 2021.

\bibitem{perceiverio}
Andrew Jaegle, Sebastian Borgeaud, Jean-Baptiste Alayrac, Carl Doersch, Catalin
  Ionescu, David Ding, Skanda Koppula, Daniel Zoran, Andrew Brock, Evan
  Shelhamer, et~al.
\newblock Perceiver io: A general architecture for structured inputs \&
  outputs.
\newblock {\em arXiv preprint arXiv:2107.14795}, 2021.

\bibitem{unit}
Ronghang Hu and Amanpreet Singh.
\newblock Unit: Multimodal multitask learning with a unified transformer.
\newblock {\em arXiv preprint arXiv:2102.10772}, 2021.

\bibitem{flava}
Amanpreet Singh, Ronghang Hu, Vedanuj Goswami, Guillaume Couairon, Wojciech
  Galuba, Marcus Rohrbach, and Douwe Kiela.
\newblock Flava: A foundational language and vision alignment model.
\newblock {\em arXiv preprint arXiv:2112.04482}, 2021.

\bibitem{uni-perceiver}
Xizhou Zhu, Jinguo Zhu, Hao Li, Xiaoshi Wu, Xiaogang Wang, Hongsheng Li,
  Xiaohua Wang, and Jifeng Dai.
\newblock Uni-perceiver: Pre-training unified architecture for generic
  perception for zero-shot and few-shot tasks.
\newblock {\em arXiv preprint arXiv:2112.01522}, 2021.

\bibitem{coatnet}
Zihang Dai, Hanxiao Liu, Quoc~V Le, and Mingxing Tan.
\newblock Coatnet: Marrying convolution and attention for all data sizes.
\newblock {\em arXiv preprint arXiv:2106.04803}, 2021.

\bibitem{bpe}
Rico Sennrich, Barry Haddow, and Alexandra Birch.
\newblock Neural machine translation of rare words with subword units.
\newblock In {\em Proceedings of the 54th Annual Meeting of the Association for
  Computational Linguistics (Volume 1: Long Papers)}, pages 1715--1725, 2016.

\bibitem{pix2seq}
Ting Chen, Saurabh Saxena, Lala Li, David~J Fleet, and Geoffrey Hinton.
\newblock Pix2seq: A language modeling framework for object detection.
\newblock {\em arXiv preprint arXiv:2109.10852}, 2021.

\bibitem{layer_norm}
Lei~Jimmy Ba, Jamie~Ryan Kiros, and Geoffrey~E. Hinton.
\newblock Layer normalization.
\newblock {\em CoRR}, abs/1607.06450, 2016.

\bibitem{normformer}
Sam Shleifer, Jason Weston, and Myle Ott.
\newblock Normformer: Improved transformer pretraining with extra
  normalization.
\newblock {\em arXiv preprint arXiv:2110.09456}, 2021.

\bibitem{tupe}
Guolin Ke, Di~He, and Tie-Yan Liu.
\newblock Rethinking positional encoding in language pre-training.
\newblock In {\em International Conference on Learning Representations}, 2020.

\bibitem{trie}
Thomas~H Cormen, Charles~E Leiserson, Ronald~L Rivest, and Clifford Stein.
\newblock {\em Introduction to algorithms}.
\newblock MIT press, 2009.

\bibitem{albef}
Junnan Li, Ramprasaath~R Selvaraju, Akhilesh~Deepak Gotmare, Shafiq Joty,
  Caiming Xiong, and Steven Hoi.
\newblock Align before fuse: Vision and language representation learning with
  momentum distillation.
\newblock In {\em Thirty-Fifth Conference on Neural Information Processing
  Systems}, 2021.

\bibitem{meter}
Zi-Yi Dou, Yichong Xu, Zhe Gan, Jianfeng Wang, Shuohang Wang, Lijuan Wang,
  Chenguang Zhu, Nanyun Peng, Zicheng Liu, and Michael Zeng.
\newblock An empirical study of training end-to-end vision-and-language
  transformers.
\newblock {\em ArXiv}, abs/2111.02387, 2021.

\bibitem{lemon}
Xiaowei Hu, Zhe Gan, Jianfeng Wang, Zhengyuan Yang, Zicheng Liu, Yumao Lu, and
  Lijuan Wang.
\newblock Scaling up vision-language pre-training for image captioning.
\newblock {\em CoRR}, abs/2111.12233, 2021.

\bibitem{Kamath2021MDETRM}
Aishwarya Kamath, Mannat Singh, Yann LeCun, Ishan Misra, Gabriel Synnaeve, and
  Nicolas Carion.
\newblock Mdetr - modulated detection for end-to-end multi-modal understanding.
\newblock {\em ArXiv}, abs/2104.12763, 2021.

\bibitem{snli-ve}
Ning Xie, Farley Lai, Derek Doran, and Asim Kadav.
\newblock Visual entailment: A novel task for fine-grained image understanding.
\newblock {\em arXiv preprint arXiv:1901.06706}, 2019.

\bibitem{coco_cap}
Xinlei Chen, Hao Fang, Tsung-Yi Lin, Ramakrishna Vedantam, Saurabh Gupta, Piotr
  Doll{\'a}r, and C~Lawrence Zitnick.
\newblock Microsoft coco captions: Data collection and evaluation server.
\newblock {\em arXiv preprint arXiv:1504.00325}, 2015.

\bibitem{refcoco}
Licheng Yu, Patrick Poirson, Shan Yang, Alexander~C Berg, and Tamara~L Berg.
\newblock Modeling context in referring expressions.
\newblock In {\em European Conference on Computer Vision}, pages 69--85.
  Springer, 2016.

\bibitem{refcocog}
Junhua Mao, Jonathan Huang, Alexander Toshev, Oana Camburu, Alan~L Yuille, and
  Kevin Murphy.
\newblock Generation and comprehension of unambiguous object descriptions.
\newblock In {\em Proceedings of the IEEE conference on computer vision and
  pattern recognition}, pages 11--20, 2016.

\bibitem{nichol2021glide}
Alex Nichol, Prafulla Dhariwal, Aditya Ramesh, Pranav Shyam, Pamela Mishkin,
  Bob McGrew, Ilya Sutskever, and Mark Chen.
\newblock Glide: Towards photorealistic image generation and editing with
  text-guided diffusion models.
\newblock {\em arXiv preprint arXiv:2112.10741}, 2021.

\bibitem{huang2021unifying}
Yupan Huang, Hongwei Xue, Bei Liu, and Yutong Lu.
\newblock Unifying multimodal transformer for bi-directional image and text
  generation.
\newblock In {\em Proceedings of the 29th ACM International Conference on
  Multimedia}, pages 1138--1147, 2021.

\bibitem{glue}
Alex Wang, Amanpreet Singh, Julian Michael, Felix Hill, Omer Levy, and Samuel~R
  Bowman.
\newblock Glue: A multi-task benchmark and analysis platform for natural
  language understanding.
\newblock {\em arXiv preprint arXiv:1804.07461}, 2018.

\bibitem{gigaword}
Alexander~M Rush, Sumit Chopra, and Jason Weston.
\newblock A neural attention model for abstractive sentence summarization.
\newblock In {\em Proceedings of the 2015 Conference on Empirical Methods in
  Natural Language Processing}, pages 379--389, 2015.

\bibitem{imagenet}
Jia Deng, Wei Dong, Richard Socher, Li-Jia Li, Kai Li, and Li~Fei-Fei.
\newblock Imagenet: A large-scale hierarchical image database.
\newblock In {\em 2009 IEEE conference on computer vision and pattern
  recognition}, pages 248--255. Ieee, 2009.

\bibitem{electra}
Kevin Clark, Minh{-}Thang Luong, Quoc~V. Le, and Christopher~D. Manning.
\newblock {ELECTRA:} pre-training text encoders as discriminators rather than
  generators.
\newblock In {\em 8th International Conference on Learning Representations,
  {ICLR} 2020}. OpenReview.net, 2020.

\bibitem{deberta}
Pengcheng He, Xiaodong Liu, Jianfeng Gao, and Weizhu Chen.
\newblock Deberta: decoding-enhanced bert with disentangled attention.
\newblock In {\em 9th International Conference on Learning Representations,
  {ICLR} 2021}. OpenReview.net, 2021.

\bibitem{rouge}
Chin-Yew Lin.
\newblock {ROUGE}: A package for automatic evaluation of summaries.
\newblock In {\em Text Summarization Branches Out}, Barcelona, Spain, July
  2004. Association for Computational Linguistics.

\bibitem{bertshare}
Sascha Rothe, Shashi Narayan, and Aliaksei Severyn.
\newblock Leveraging pre-trained checkpoints for sequence generation tasks.
\newblock {\em Transactions of the Association for Computational Linguistics},
  8:264--280, 2020.

\bibitem{mass}
Kaitao Song, Xu~Tan, Tao Qin, Jianfeng Lu, and Tie{-}Yan Liu.
\newblock {MASS:} masked sequence to sequence pre-training for language
  generation.
\newblock In {\em {ICML} 2019}, pages 5926--5936, 2019.

\bibitem{pegasus}
Jingqing Zhang, Yao Zhao, Mohammad Saleh, and Peter Liu.
\newblock Pegasus: Pre-training with extracted gap-sentences for abstractive
  summarization.
\newblock In {\em International Conference on Machine Learning}, pages
  11328--11339. PMLR, 2020.

\bibitem{prophetnet}
Weizhen Qi, Yu~Yan, Yeyun Gong, Dayiheng Liu, Nan Duan, Jiusheng Chen, Ruofei
  Zhang, and Ming Zhou.
\newblock Prophetnet: Predicting future n-gram for sequence-to-sequence
  pre-training.
\newblock In {\em Proceedings of the 2020 Conference on Empirical Methods in
  Natural Language Processing: Findings}, pages 2401--2410, 2020.

\bibitem{tan2019efficientnet}
Mingxing Tan and Quoc Le.
\newblock Efficientnet: Rethinking model scaling for convolutional neural
  networks.
\newblock In {\em International Conference on Machine Learning}, pages
  6105--6114. PMLR, 2019.

\bibitem{dino}
Mathilde Caron, Hugo Touvron, Ishan Misra, Herv{\'e} J{\'e}gou, Julien Mairal,
  Piotr Bojanowski, and Armand Joulin.
\newblock Emerging properties in self-supervised vision transformers.
\newblock {\em arXiv preprint arXiv:2104.14294}, 2021.

\bibitem{cc12m}
Soravit Changpinyo, Piyush Sharma, Nan Ding, and Radu Soricut.
\newblock Conceptual 12m: Pushing web-scale image-text pre-training to
  recognize long-tail visual concepts.
\newblock In {\em Proceedings of the IEEE/CVF Conference on Computer Vision and
  Pattern Recognition}, pages 3558--3568, 2021.

\bibitem{cc}
Piyush Sharma, Nan Ding, Sebastian Goodman, and Radu Soricut.
\newblock Conceptual captions: {A} cleaned, hypernymed, image alt-text dataset
  for automatic image captioning.
\newblock In {\em {ACL} 2018}, pages 2556--2565, 2018.

\bibitem{sbu}
Vicente Ordonez, Girish Kulkarni, and Tamara~L. Berg.
\newblock Im2text: Describing images using 1 million captioned photographs.
\newblock In {\em NeurIPS 2011}, pages 1143--1151, 2011.

\bibitem{vg}
Ranjay Krishna, Yuke Zhu, Oliver Groth, Justin Johnson, Kenji Hata, Joshua
  Kravitz, Stephanie Chen, Yannis Kalantidis, Li{-}Jia Li, David~A. Shamma,
  Michael~S. Bernstein, and Li~Fei{-}Fei.
\newblock Visual genome: Connecting language and vision using crowdsourced
  dense image annotations.
\newblock {\em International Journal of Computer Vision}, 123(1):32--73, 2017.

\bibitem{vqav2}
Yash Goyal, Tejas Khot, Douglas Summers-Stay, Dhruv Batra, and Devi Parikh.
\newblock Making the v in vqa matter: Elevating the role of image understanding
  in visual question answering.
\newblock In {\em Proceedings of the IEEE Conference on Computer Vision and
  Pattern Recognition}, pages 6904--6913, 2017.

\bibitem{gqa}
Drew~A Hudson and Christopher~D Manning.
\newblock Gqa: A new dataset for real-world visual reasoning and compositional
  question answering.
\newblock In {\em CVPR 2019}, pages 6700--6709, 2019.

\bibitem{yfcc100m}
Bart Thomee, David~A Shamma, Gerald Friedland, Benjamin Elizalde, Karl Ni,
  Douglas Poland, Damian Borth, and Li-Jia Li.
\newblock Yfcc100m: The new data in multimedia research.
\newblock {\em Communications of the ACM}, 59(2):64--73, 2016.

\bibitem{openimages}
Alina Kuznetsova, Hassan Rom, Neil Alldrin, Jasper Uijlings, Ivan Krasin, Jordi
  Pont-Tuset, Shahab Kamali, Stefan Popov, Matteo Malloci, Alexander
  Kolesnikov, et~al.
\newblock The open images dataset v4.
\newblock {\em International Journal of Computer Vision}, 128(7):1956--1981,
  2020.

\bibitem{object365}
Shuai Shao, Zeming Li, Tianyuan Zhang, Chao Peng, Gang Yu, Xiangyu Zhang, Jing
  Li, and Jian Sun.
\newblock Objects365: A large-scale, high-quality dataset for object detection.
\newblock In {\em Proceedings of the IEEE/CVF International Conference on
  Computer Vision}, pages 8430--8439, 2019.

\bibitem{pile}
Leo Gao, Stella Biderman, Sid Black, Laurence Golding, Travis Hoppe, Charles
  Foster, Jason Phang, Horace He, Anish Thite, Noa Nabeshima, et~al.
\newblock The pile: An 800gb dataset of diverse text for language modeling.
\newblock {\em arXiv preprint arXiv:2101.00027}, 2020.

\bibitem{resnet}
Kaiming He, Xiangyu Zhang, Shaoqing Ren, and Jian Sun.
\newblock Deep residual learning for image recognition.
\newblock In {\em {CVPR} 2016}, pages 770--778, 2016.

\bibitem{adamw}
Ilya Loshchilov and Frank Hutter.
\newblock Decoupled weight decay regularization.
\newblock In {\em {ICLR} 2019}, 2019.

\bibitem{drop-path}
Gao Huang, Yu~Sun, Zhuang Liu, Daniel Sedra, and Kilian~Q. Weinberger.
\newblock Deep networks with stochastic depth.
\newblock In {\em ECCV}, 2016.

\bibitem{bleu}
Kishore Papineni, Salim Roukos, Todd Ward, and Wei-Jing Zhu.
\newblock Bleu: a method for automatic evaluation of machine translation.
\newblock In {\em Proceedings of the 40th annual meeting of the Association for
  Computational Linguistics}, pages 311--318, 2002.

\bibitem{meteor}
Satanjeev Banerjee and Alon Lavie.
\newblock Meteor: An automatic metric for mt evaluation with improved
  correlation with human judgments.
\newblock In {\em Proceedings of the acl workshop on intrinsic and extrinsic
  evaluation measures for machine translation and/or summarization}, pages
  65--72, 2005.

\bibitem{cider}
Ramakrishna Vedantam, C~Lawrence~Zitnick, and Devi Parikh.
\newblock Cider: Consensus-based image description evaluation.
\newblock In {\em Proceedings of the IEEE conference on computer vision and
  pattern recognition}, pages 4566--4575, 2015.

\bibitem{spice}
Peter Anderson, Basura Fernando, Mark Johnson, and Stephen Gould.
\newblock Spice: Semantic propositional image caption evaluation.
\newblock In {\em European conference on computer vision}, pages 382--398.
  Springer, 2016.

\bibitem{karpathy}
Andrej Karpathy and Li~Fei-Fei.
\newblock Deep visual-semantic alignments for generating image descriptions.
\newblock In {\em Proceedings of the IEEE conference on computer vision and
  pattern recognition}, pages 3128--3137, 2015.

\bibitem{heusel2017gans}
Martin Heusel, Hubert Ramsauer, Thomas Unterthiner, Bernhard Nessler, and Sepp
  Hochreiter.
\newblock Gans trained by a two time-scale update rule converge to a local nash
  equilibrium.
\newblock {\em Advances in neural information processing systems}, 30, 2017.

\bibitem{salimans2016improved}
Tim Salimans, Ian Goodfellow, Wojciech Zaremba, Vicki Cheung, Alec Radford, and
  Xi~Chen.
\newblock Improved techniques for training gans.
\newblock {\em Advances in neural information processing systems},
  29:2234--2242, 2016.

\bibitem{rennie2017self}
Steven~J Rennie, Etienne Marcheret, Youssef Mroueh, Jerret Ross, and Vaibhava
  Goel.
\newblock Self-critical sequence training for image captioning.
\newblock In {\em Proceedings of the IEEE conference on computer vision and
  pattern recognition}, pages 7008--7024, 2017.

\bibitem{el}
Guangxiang Zhao, Wenkai Yang, Xuancheng Ren, Lei Li, and Xu~Sun.
\newblock Well-classified examples are underestimated in classification with
  deep neural networks.
\newblock {\em CoRR}, abs/2110.06537, 2021.

\bibitem{rand_augment}
Ekin~D Cubuk, Barret Zoph, Jonathon Shlens, and Quoc~V Le.
\newblock Randaugment: Practical automated data augmentation with a reduced
  search space.
\newblock In {\em Proceedings of the IEEE/CVF Conference on Computer Vision and
  Pattern Recognition Workshops}, pages 702--703, 2020.

\bibitem{random_erase}
Zhun Zhong, Liang Zheng, Guoliang Kang, Shaozi Li, and Yi~Yang.
\newblock Random erasing data augmentation.
\newblock In {\em Proceedings of the AAAI Conference on Artificial
  Intelligence}, volume~34, pages 13001--13008, 2020.

\bibitem{mixup}
Hongyi Zhang, Moustapha Ciss{\'{e}}, Yann~N. Dauphin, and David Lopez{-}Paz.
\newblock mixup: Beyond empirical risk minimization.
\newblock In {\em 6th International Conference on Learning Representations,
  {ICLR} 2018, Vancouver, BC, Canada, April 30 - May 3, 2018, Conference Track
  Proceedings}. OpenReview.net, 2018.

\bibitem{cutmix}
Sangdoo Yun, Dongyoon Han, Sanghyuk Chun, Seong~Joon Oh, Youngjoon Yoo, and
  Junsuk Choe.
\newblock Cutmix: Regularization strategy to train strong classifiers with
  localizable features.
\newblock In {\em 2019 {IEEE/CVF} International Conference on Computer Vision,
  {ICCV} 2019, Seoul, Korea (South), October 27 - November 2, 2019}, pages
  6022--6031. {IEEE}, 2019.

\bibitem{schuhmann2021laion}
Christoph Schuhmann, Richard Vencu, Romain Beaumont, Robert Kaczmarczyk,
  Clayton Mullis, Aarush Katta, Theo Coombes, Jenia Jitsev, and Aran
  Komatsuzaki.
\newblock Laion-400m: Open dataset of clip-filtered 400 million image-text
  pairs.
\newblock {\em arXiv preprint arXiv:2111.02114}, 2021.

\end{thebibliography}

\newpage
\appendix
\section{Implementation Details}
\label{app:implementation_details}

\subsection{Pretraining Datasets}
\label{sec:appendix_pretraining_datasets}
We construct pretraining datasets by incorporating Vision \& Language data (i.e., image-text pairs), Vision data (i.e., raw image data, object-labeled data), and Language data (i.e., plain texts). 
For replication, the pretraining datasets are publicly available. 
We carefully filter our pretraining data and exclude images that appear in the validation and test sets of downstream tasks to avoid data leakage. 
The statistics on the pretraining datasets are listed in Table \ref{tb:datasets-v2}. 

\paragraph{Cross-modal Data}
For vision \& language pretraining, we mainly apply image-text pairs, including image-caption pairs, image-QA pairs, and image-region pairs, as the pretraining data. 
For the pretraining tasks of image captioning and image-text matching, we collect Conceptual Caption 12M (CC12M)~\cite{cc12m}, Conceptual Captions (CC3M)~\cite{cc}, SBU~\cite{sbu}, MSCOCO image captions (COCO)~\cite{coco_cap}, and Visual Genome Captions (VG Captions)~\cite{vg}. 
Specifically, the part of data from VG requires some additional processing. As texts in VG captions describe local regions on the images, we retrieve regions with area larger than $16,384$ pixels and construct region-caption pairs.
For visual question answering, we collect VQAv2~\cite{vqav2}, VG-QA~\cite{vg}, as well as GQA~\cite{gqa}. VQAv2 is a visual question answering dataset with real-world photographs from COCO. VG-QA is also a visual question answering dataset with real-world photographs from VG. The questions of VG-QA are related to specific regions on the images. GQA is a large VQA dataset featuring compositional questions. The images of GQA are also collected from VG.
For visual grounding and grounded captioning, we collect data from RefCOCO~\cite{refcoco}, RefCOCO+~\cite{refcoco}, RefCOCOg~\cite{refcocog} and VG captions. Additional processing is applied to VG Captions for this task. Specifically, we use the data of VG that contains regions with area smaller than $16,384$ pixels for Visual Grounding, in order to encourage model to grasp fine-grained alignments between vision and language.

\paragraph{Uni-modal Data}
Uni-modal data includes vision and language data. 
Vision data consists of raw images for image infilling and object-labeled images for object detection. 
For image infilling, we collect raw images from OpenImages, YFCC100M~\cite{yfcc100m} and ImageNet-21K~\cite{imagenet}, and exclude annotations. Thus the model is unable to access labels in the pretraining stage. 
For object detection, we collect OpenImages~\cite{openimages}, Object365~\cite{object365}, VG and COCO for object detection. 
Language data consists of plain texts, i.e., passages consisting of sentences. We use around 140GB of data from Pile~\cite{pile} to leverage its diversity. Specifically, we extract natural language data and implement preprocessing methods, including truncation to the length of $512$.

\begin{table*}[h]
\centering
\caption{Statistics on the datasets of pretraining tasks. ``\#Image'' denotes the number of distinct images, and ``\#Sample'' denotes the number of samples. *For language data, we report its storage following the previous studies~\cite{bert, roberta}.}
% For language data, 140GB represents the storage space of the plain texts.}
\vskip 0.15in
\begin{adjustbox}{max width=1.\textwidth}
\begin{tabular}{@{\extracolsep{\fill}}ccccccc}
\toprule
  Type
  &Pretraining Task
  &Source
  &\#Image
  &\#Sample
  \\
\midrule
  \multirow{5}*{Vision \& Language}
  &Image Captioning
  &\multirow{2}*{CC12M, CC3M, SBU, COCO, VG-Cap}
  &\multirow{2}*{14.78M}
  &\multirow{2}*{15.25M}
  \\
  &Image-Text Matching
  \\
\cmidrule{2-5}
  &Visual Question Answering
  &VQAv2, VG-QA, GQA
  &178K
  &2.92M
  \\
\cmidrule{2-5}
  &Visual Grounding
  &\multirow{2}*{RefCOCO, RefCOCO+, RefCOCOg, VG-Cap}
  &\multirow{2}*{131K}
  &\multirow{2}*{3.20M}
  \\
  &Grounded Captioning
  \\
\midrule
  \multirow{2}*{Vision}
  &Detection
  &OpenImages, Object365, VG, COCO
  &2.98M
  &3.00M
  \\
\cmidrule{2-5}
  &Image Infilling
  &OpenImages, YFCC100M, ImageNet-21K
  &36.27M
  &-
  \\
\midrule
  Language
  &Masked Language Modeling
  &Pile (Filtered)
  &-
  &140GB*
  \\
\bottomrule
\end{tabular}
\end{adjustbox}
\label{tb:datasets-v2}
\end{table*}

\subsection{Pretraining Details}
\label{sec:appendix_pretrain_details}

% Our network configuration is similar to BART~\cite{bart}. $\rm \ofa_{Base}$ consists of $6$ encoder layers and $6$ decoder layers, with the hidden size $768$ and $12$ attention heads in each layer. $\rm \ofa_{Large}$ consists of 12 encoder layers and 12 decoder layers, with the hidden size $1,024$ and $16$ attention heads in each layer. The intermediate sizes of FFN are 3072 and 4096 for \textit{Base} and \textit{Large} models, respectively.

For the image processing, we first resize and crop the images into different resolutions, $256 \times 256$ for $\text{OFA}\rm_{Tiny}$ and $\text{OFA}\rm_{Medium}$, $384 \times 384$ for $\text{OFA}\rm_{Base}$, $480 \times 480$ for $\text{OFA}\rm_{Large}$ and $\text{OFA}\rm_{Huge}$,  with a fixed patch size of $16 \times 16$. 
Note that training $\text{OFA}\rm_{Large}$ and $\text{OFA}\rm_{Huge}$ are time and computation consuming, we first train them with images of the resolution of $384 \times 384$ and $256 \times 256$, and continue pretraining with images of the resolution of $480 \times 480$. 

For each patch, we obtain its feature vector with the first three blocks of ResNet~\cite{resnet}. 
The ResNet module is jointly trained along with the transformer module. 
Note that through extensive experiments we find that random sampling patches~\cite{pixelbert} does not bring additional benefits in our scenario. 
% Following~\cite{pixelbert}, we randomly sample $196$ patch features to improve the robustness of feature learning during the pretraining stage.
For the text processing, we tokenize the texts with the same BPE Tokenizer~\cite{bpe} as BART~\cite{bart}. 
The maximum text sequence length of both encoder and decoder is set to $256$. 
We share parameters between the embedding and the decoder softmax output layer. 

From our preliminary experiments, we find that the initialization for Transformer plays an important role. For $\text{OFA}\rm_{Base}$ and $\text{OFA}\rm_{Large}$, we initialize the transformer with most of the weights of $\text{BART}\rm_{Base}$ and $\text{BART}\rm_{Large}$ considering the slight difference between OFA Transformer and BART as described in Sec~\ref{sec:unified_io_arc}. For OFA of the other sizes, we pretrain language models with the same pretraining strategy with BART and use the pretrained weights to initialize the Transformer in OFA. 

We use the AdamW~\cite{adamw} optimizer with $(\beta_1,\beta_2)=(0.9, 0.999)$ and $\epsilon=1e\text{-}8$ to pretrain our models. We set the peak learning rate to $2e\text{-}4$, and apply a scheduler with linear decay with a warmup ratio of $0.01$ to control the learning rate. 
For regulation, we set dropout to $0.1$ and use weight decay with $0.01$. We employ stochastic depth~\cite{drop-path} with a $0.1$ rate (applied to encoder and decoder except for convolution blocks).
We mix all the pretraining data within each batch, which contains $2,048$ vision\&language samples, $256$ object detection samples, $256$ image-only samples and $512$ text-only samples. 
All models are pretrained for at least $300K$ steps except the models used for ablation study. 
% Both  model are pretrained for $500K$ steps and $250K$ steps, respectively.

\subsection{Details of Downstream Tasks}
\label{sec:appendix_downstream}

We verify the capability of \ofa~on various downstream tasks in both finetuning and zero-shot settings. 
We design various task-specific instructions to transfer the knowledge learned from pretraining to downstream tasks effectively. The instructions of different tasks are listed in Table~\ref{tb:downstream-task}. 
For finetuning, if not specified, the input image resolution is set to $480 \times 480$, and the other hyper-parameters remain the same as for pretraining.
% For finetuning, we use the AdamW optimizer with the same beta values as used in the pretraining stage.
% For regulation, we set dropout to $0.1$ and apply stochastic depth~\cite{drop-path} to the encoder and decoder except for convolution blocks.
% If not specified, the stochastic depth rate is set to $0.1$ and the input image resolution is set to $480 \times 480$.
% Following standard practice, we choose the hyper-parameters for each task based on the model performance on the development set.
The experimental details of different downstream tasks, including both multimodal and uni-modal tasks, are listed below:

\paragraph{Image Captioning} 
Image captioning is a standard vision\&language task that requires models to generate an appropriate and fluent caption for an image.
We adopt the most widely used MSCOCO Image Caption dataset~\cite{coco_cap} to evaluate the multi-modal generation capability of \ofa. We report BLEU-4~\cite{bleu}, METEOR~\cite{meteor}, CIDEr~\cite{cider}, and SPICE~\cite{spice} scores on the Karpathy test split~\cite{karpathy}.
Following the previous standard practice, we first finetune \ofa~with cross-entropy loss for $2$ epochs with a batch size of $128$ and a learning rate of $1e-5$, and label smoothing is set to $0.1$. We then finetune the model with CIDEr optimization for $3$ epochs with a batch size of $64$, and disable dropout and stochastic depth. We report both scores at the two stages.

\paragraph{Visual Question Answering}
Visual question answering (VQA) is a cross-modal task that requires the models to answer the question given an image. 
Previous works such as VLMo~\cite{vlmo} or SimVLM~\cite{simvlm} define VQA as a classification task. They use a linear output layer to predict the probability of each candidate answer on a given set. 
In contrast with these studies, to adapt the generative \ofa~model to VQA benchmark, we use the Trie-based search strategy mentioned in Sec.~\ref{sec:finetune_inference} to ensure that the answer generated by \ofa~is constrained in the candidate set. 
We evaluate our model with other baselines on the commonly used VQAv2 dataset~\cite{vqav2}.
Accuracy scores on both test-dev and test-std sets are reported. The \ofa~models of all the reported sizes are finetuned for $40,000$ steps with a batch size of $512$. The learning rate is $5e-5$ with the label smoothing of $0.1$.
When finetuning $\text{OFA}\rm_{Large}$ and $\text{OFA}\rm_{Huge}$, we increase the image resolution from $480$ to $640$. Linear interpolation of the image absolute positional embedding proposed in \cite{vit} is employed when transferring the pretrained $\text{OFA}$ to VQA finetuning.
During Trie-based searching, we constrain the generated answers over the most frequent $3,129$ answer candidates. Exponential moving average (EMA) with decay rate $0.9999$ is employed in finetuning.

% \begin{table*}[t]
% \small
% \centering
% \begin{adjustbox}{}
% \begin{tabular}{@{\extracolsep{\fill}}l|l|l|}
% \toprule
%   Downstream Task
%   &Instruction
%   &Target 
%   \\
% \midrule
%   COLA
%   &Is the sentence ``\{\textbf{Sentence}\}" grammatically correct?
%   &Yes / No
%   \\
%   SST-2
%   &Is the sentiment of sentence ``\{\textbf{Sentence}\}" positive or negative?
%   &Positive / Negative
%   \\
%   RTE
%   &Can sentence ``\{\textbf{Sentence1}\}" imply sentence ``\{\textbf{Sentence2}\}"?
%   &Yes / No
%   \\
%   MRPC
%   &Does sentence1 ``\{\textbf{Sentence1}\}" and sentence2 ``\{\textbf{Sentence2}\}" have the same semantics?
%   &Yes / No
%   \\
%   QQP
%   &Is question ``\{\textbf{Question1}\}" and question ``\{\textbf{Question2}\}" equivalent?
%   &Yes / No
%   \\
%   MNLI
%   &Can sentence ``\{\textbf{Sentence1}\}" be derived from sentence ``\{\textbf{Sentence2}\}"?
%   &Yes / No / Maybe
%   \\
%   QNLI
%   &Does sentence ``\{\textbf{Sentence}\}" contain the answer to question ``\{\textbf{Question}\}"?
%   &Yes / No
%   \\
%   WNLI
%   &Can sentence ``\{\textbf{Sentence1}\}" imply sentence ``\{\textbf{Sentence2}\}"?
%   &Yes / No
%   \\
%   Text Summarization
%   &What is the summary of article ``\{\textbf{Article}\}"?
%   &\{\textbf{Summary}\}
%   \\
% \bottomrule
% \end{tabular}
% \end{adjustbox}
% \caption{Instructions for downstream tasks.}
% \label{tb:downstream-task}
% \end{table*}

\begin{table*}[t]
\caption{Instructions for downstream tasks.}
\vskip 0.15in
\centering
\begin{adjustbox}{max width=1.\textwidth}
\begin{tabular}{@{\extracolsep{\fill}}cccc}
\toprule
Task &
  Dataset &
  Instruction &
  Target \\ \midrule
Image Captioning &
  COCO &
  $\rm \left[\textbf{Image}\right]$ What does the image describe? &
  \{\textbf{Caption}\} \\ \midrule
  \begin{tabular}[c]{@{}c@{}}Visual Question \\ Answering\end{tabular} &
  VQA &
  $\rm \left[\textbf{Image}\right]$ \{\textbf{Question}\} &
  \{\textbf{Answer}\} \\ \midrule
Visual Entailment &
  SNLI-VE &
  $\rm \left[\textbf{Image}\right]$ Can image and text1 ``\{\textbf{Text1}\}" imply text2 ``\{\textbf{Text2}\}"? &
  Yes/No/Maybe \\ \midrule
\begin{tabular}[c]{@{}c@{}}Referring Expression \\ Comprehension\end{tabular} &
  \begin{tabular}[c]{@{}c@{}}RefCOCO, \\ RefCOCO+, \\ RefCOCOg\end{tabular} &
  $\rm \left[\textbf{Image}\right]$ Which region does the text ``\{\textbf{Text}\}" describe? &
  \begin{tabular}[c]{@{}c@{}}\{\textbf{Location}\}\end{tabular} 
\\ \midrule
Image Generation &
  COCO &
  What is the complete image? caption: \{\textbf{Caption}\} &
  \{\textbf{Image}\} \\ \midrule
Image Classification &
  ImageNet-1K &
  $\rm \left[\textbf{Image}\right]$ What does the image describe? &
  \{\textbf{Label}\} \\ \midrule
  \begin{tabular}[c]{@{}c@{}}Single-Sentence \\ Classification\end{tabular} &
  SST-2 &
  Is the sentiment of text ``\{\textbf{Text}\}" positive or negative? &
  Positive/Negative \\ \midrule
\multirow{6}{*}{\begin{tabular}[c]{@{}c@{}}Sentence-Pair \\ Classification\end{tabular}} &
  RTE &
  Can text1 ``\{\textbf{Text1}\}" imply text2 ``\{\textbf{Text2}\}"? &
  Yes/No \\
 &
  MRPC &
  Does text1 ``\{\textbf{Text1}\}" and text2 ``\{\textbf{Text2}\}" have the same semantics? &
  Yes/No \\
 &
  QQP &
  Is question ``\{\textbf{Question1}\}" and question ``\{\textbf{Question2}\}" equivalent? &
  Yes/No \\
 &
  MNLI &
  Can text1 ``\{\textbf{Text1}\}" imply text2 ``\{\textbf{Text2}\}"? &
  Yes/No/Maybe \\
 &
  QNLI &
  Does ``\{\textbf{Text}\}" contain the answer to question ``\{\textbf{Question}\}"? &
  Yes/No \\ \midrule
Text Summarization &
  Gigaword &
  What is the summary of article ``\{\textbf{Article}\}"? &
  \{\textbf{Summary}\} \\
\bottomrule
\end{tabular}
\end{adjustbox}

\label{tb:downstream-task}
\end{table*}

\paragraph{Visual Entailment} 
Visual entailment requires the model to evaluate how the given image and text are semantically correlated, i.e., entailment, neutral, or contradiction. We perform experiments on the SNLI-VE dataset~\cite{snli-ve}. The image premise, text premise and text hypothesis are fed to the encoder, and the decoder generates appropriate labels. To transfer the knowledge learned by pretraining to this task, we convert the labels entailment/neutral/contradiction to yes/maybe/no. 
We also use the Trie-based search strategy to constrain the generated labels over the candidate set.
We report accuracy on both dev and test sets. 
The \ofa~model is finetuned for $6$ epochs with a learning rate of $2e-5$ and a batch size of $256$. 
% The stochastic depth rate is set to $0.2$.

\paragraph{Referring Expression Comprehension} 
Referring expression comprehension requires models to locate an image region described by a language query. Different from the approach taken by most previous methods~\cite{vilbert,uniter} which ranks a set of candidate bounding boxes detected by a pretrained object detector, our method directly predicts the best matching bounding box without any proposals. We perform experiments on RefCOCO~\cite{refcoco}, RefCOCO+~\cite{refcoco}, and RefCOCOg~\cite{refcocog}. Consistent with other downstream tasks, we formulate referring expression comprehension as a conditional
sequence generation task. In detail, given an image and a language query, \ofa~generates the box sequence (e.g., $\langle x_1, y_1, x_2, y_2 \rangle$) in an autoregressive manner. We report the standard metric Acc@0.5 on the validation and test sets.
For finetuning, the input image resolution is set to $512\times 512$.
We finetune the \ofa~model on each dataset for about $10$ epochs with a batch size of $128$. The learning rate is $3e-5$ with the label smoothing of $0.1$.
Each query only corresponds to an image region, so we limit the maximum generated length to $4$ during inference.

\paragraph{Image Generation}
Following the same setting with~\cite{nvwa}, we train our model on the MS COCO train split and evaluate our model on the validation split by randomly sampling $30,000$ images.
We use Fréchet Inception Distance (FID) \cite{heusel2017gans} and Inception Score (IS) \cite{salimans2016improved} to evaluate the quality of the images. 
Following the previous studies \cite{huang2021unifying, nvwa}, we also compute CLIP Similarity Score (CLIPSIM) to evaluate the semantic similarity between the query text and the generated images.
During finetuning, \ofa~learns to generate the image code sequence according to the given text query only.
The model is first finetuned with cross-entropy and then with CLIPSIM optimization following~\cite{rennie2017self,huang2021unifying}.
In the first stage, we finetune the \ofa~ model for about $50$ epochs with a batch size of $512$ and a learning rate of $1e-3$. In the second stage, the model is finetuned for extra $5000$ steps with a batch size of $32$ and a learning rate of $1e-6$.
During the evaluation, we sample $24$ images with the resolution of $256 \times 256$ for each query and choose the best one using the pretrained CLIP model~\cite{clip}. 

For case study, we compare \ofa~with CogView and GLIDE. CogView provides an API website~\footnote{\url{https://wudao.aminer.cn/CogView/index.html}}. 
Note that this API samples 8 images of resolution of $512 \times 512$ for each query. We select the first one of generated images and resize it to the resolution of $256 \times 256$. 
GLIDE provides a Colab notebook.\footnote{\url{https://colab.research.google.com/drive/1q6tJ58UKod1eCOkbaUNGzF3K5BbXlB5m}}. 
Note that the only publicly available GLIDE model is of \textit{base} size ($\sim$385M). 

\paragraph{Image Classification}
We provide finetuning results on ImageNet-1K~\cite{imagenet} following recent studies in self-supervised learning for computer vision. During finetuning and inference, a Trie-based search strategy is employed to constrain the generated text into the set of $1,000$ candidate labels. We finetune \ofa~for $32$ epochs and a batch size of $256$. The learning rate is $5e-5$. The ratio for label smoothing is $0.1$. The encouraging loss proposed in \cite{el} is employed with the hyperparameter LE set to $0.75$. Following \cite{beit}, we use the same random resize cropping, random flipping, RandAug~\cite{rand_augment} and random erasing~\cite{random_erase} transformations as data augmentation strategies. Mixup~\cite{mixup} and CutMix~\cite{cutmix} are used with overall $0.5$ probability to be performed on each batch and alpha is $0.8$ and $1.0$, respectively. To adapt the mixed soft target of Mixup and CutMix into generation paradigm during finetuning, we run the decoder twice each with one of the target sequences to be mixed and sum the loss weighted by the mixing ratio.

\paragraph{Natural Language Understanding}
To verify the natural language understanding ability of \ofa, we select $6$ language understanding tasks from GLUE benchmark~\cite{glue}, including both single-sentence classification tasks and sentence-pair classification tasks. To adapt to sentence-pair classification, previous models~\cite{bert, roberta} usually use segment embeddings to distinguish different sentences. Unlike those models, \ofa~can apply the model to sentence-pair classification tasks by constructing appropriate instructions without introducing additional segment embeddings.
For the hyper-parameters of finetuning, we tune the training epochs among $\{5, 7, 10\}$, learning rate among $\{3e-5, 5e-5, 6e-5, 7e-5, 1e-4\}$, batch size among $\{32, 64, 128\}$, weight decay among $\{0.01, 0.05\}$, and dropout rate among $\{0.0, 0.1\}$. We report the best performance on the development set for each task.

\paragraph{Natural Language Generation}
We verify the natural language generation ability of \ofa~in the Gigaword dataset~\cite{gigaword}. We report ROUGE-1/ROUGE-2/ROUGE-L to evaluate the generation results following \cite{gigaword}. 
We finetune the \ofa~models for $6$ epochs with a batch size of $512$. The learning rate is $1e-4$ with the label smoothing of $0.1$, and the maximum input text sequence length is set to $512$. 
During inference, we set the length penalty to $0.7$ and beam size to $6$, and limit the maximum generated length to 32.

\section{Trie-based Search}
\label{sec:trie_search}
\begin{figure}[ht!]
\vskip 0.2in
    \centering
    \includegraphics[width=0.75\linewidth]{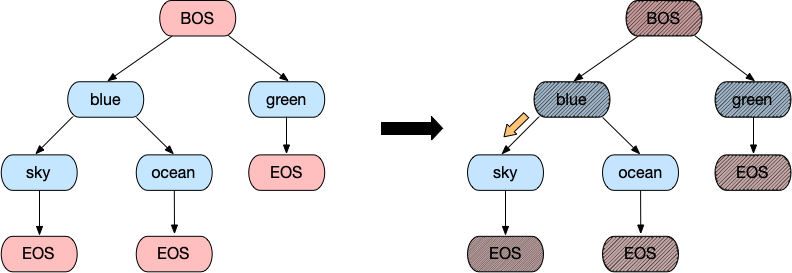}
    \caption{Example of Trie-based search where the constraint labels are ``blue sky'', ``blue ocean'' and ``green''. When computing the log-prob of token ``sky'', we only consider tokens in \{``sky'', ``ocean''\} and forcefully set the logits for all invalid tokens to $-\infty$.}
    \label{fig:trie}
% \vskip -0.2in
\end{figure}

This section describes how to use Trie-based search to improve model performance on downstream classification tasks.
When dealing with classification tasks, we first construct a Trie where nodes are annotated with tokens from the candidate label-set. 
% The model is forced to only consider tokens in the Trie.
During finetuning, the model computes the log-probabilities of the target tokens based on their positions on the Trie.
As shown in Figure~\ref{fig:trie}, when computing the log-probabilities of the target token ``sky'', we only consider tokens in \{``sky'', ``ocean''\} and forcefully set the logits for all invalid tokens to $-\infty$.
During inference, we constrain the generated labels over the candidate set.
As shown in Table~\ref{tb:trie-results}, Trie-based search strategy can boost the performance of \ofa~in various downstream classification tasks.

\begin{table}[ht]
\caption{Ablation results of Trie. The removal of Trie-based search degenerates the performance on downstream tasks. Note that the baseline $\text{OFA}\rm_{Base}$ is only pre-trained for 250k steps, which is also used in Table~\ref{tb:ablation-results}.}
\vskip 0.15in
\centering
\small
\begin{adjustbox}{max width=1.\textwidth}
\begin{tabular}{@{\extracolsep{\fill}}lccccc}
\toprule
  \multirow{2}*{Model}
  &VQA
  &SNLI-VE
  &ImageNet
  &MRPC
  &QQP
  \\
  &Test-dev Acc.
  &Dev Acc.
  &Top-1 Acc.
  &F1
  &F1
  \\
\midrule
  $\text{OFA}\rm_{Base}$
  &76.03
  &89.2
  &82.2
  &90.6
  &88.4
  \\
  \ \ \textit{w/o Trie}
  &75.86(-0.17)
  &89.0(-0.2)
  &81.9(-0.3)
  &90.1(-0.5)
  &88.2(-0.2)
  \\
\bottomrule
\end{tabular}
\end{adjustbox}
\label{tb:trie-results}
\end{table}

% \begin{figure*}[htbp]
% \centering
% \begin{minipage}[ht!]{.45\linewidth}
%  \centering
%  \includegraphics[width=0.6\linewidth]{pic/trie.png}
%  \label{tb:zero-shot-glue-results}
% \end{minipage}
% \begin{minipage}[ht!]{.45\linewidth}
%  \centering
%  \begin{tabular}{@{\extracolsep{\fill}}ccc}
%  \toprule
 
%   &OFA
%   &OFA(w/o Trie)
%   \\
%   \midrule
%   VQA
%   &76.03
%   &75.86(-0.17)
%   \\
%   ImageNet
%   &82.2
%   &81.9(-0.3)
%   \\
%  \bottomrule
%  \end{tabular}
% \end{minipage}
% \end{figure*}

\section{Qualitative Examples}
\label{sec:qualitative_examples}
This section provides more qualitative examples of multiple tasks, including text-to-image generation, open-domain VQA, grounded question answering, and open-domain visual grounding, from the generation of \ofa. By reading this section, we hope that readers can better perceive~\ofa.

% \input{table/finetuning_details}
% \input{table/datasets_v2}
% \input{table/zero_shot_nlu_result}

% \begin{figure*}[t]
%     \centering
%     \includegraphics[width=1.0\linewidth]{pic/downstream_v2.pdf}
%     \caption{Instructions for cross-modal downstream tasks }
%     \label{fig:downstream}
% \end{figure*}
% \input{table/finetuning_details}

\begin{figure*}[t]
    \centering
    \includegraphics[width=1.0\linewidth]{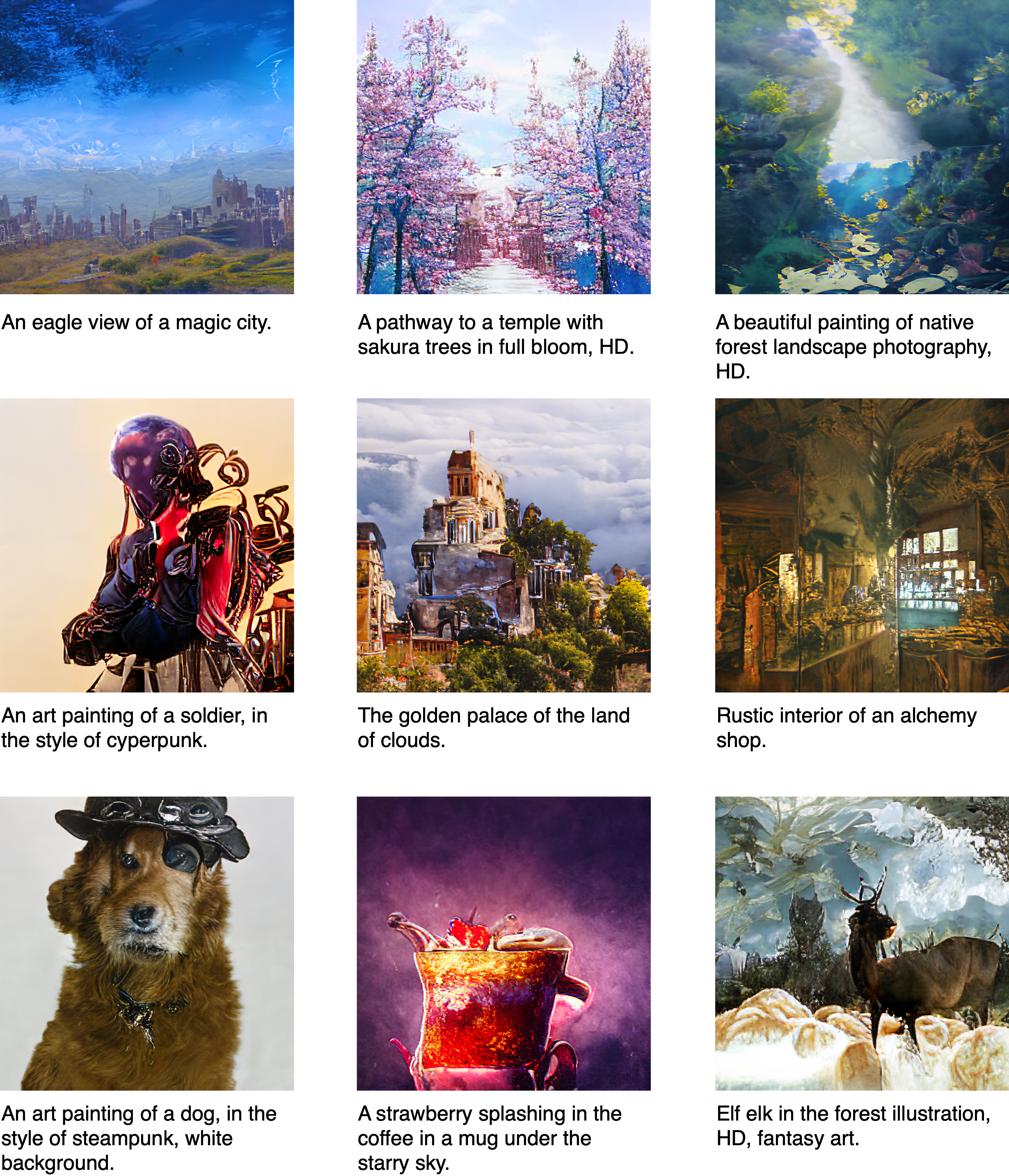}
    \caption{Examples of text-to-image generation. For better demonstration, we continue finetuning OFA on a subset of LAION-400M~\cite{schuhmann2021laion}.}
    % In this demonstration, we continue finetuning on a subset of LAION-400M~\cite{schuhmann2021laion} to further exploit the potential of the model for this task. }
    \label{fig:image_gen_samples1}
\end{figure*}

\begin{figure*}[t]
    \centering
    \includegraphics[width=1.0\linewidth]{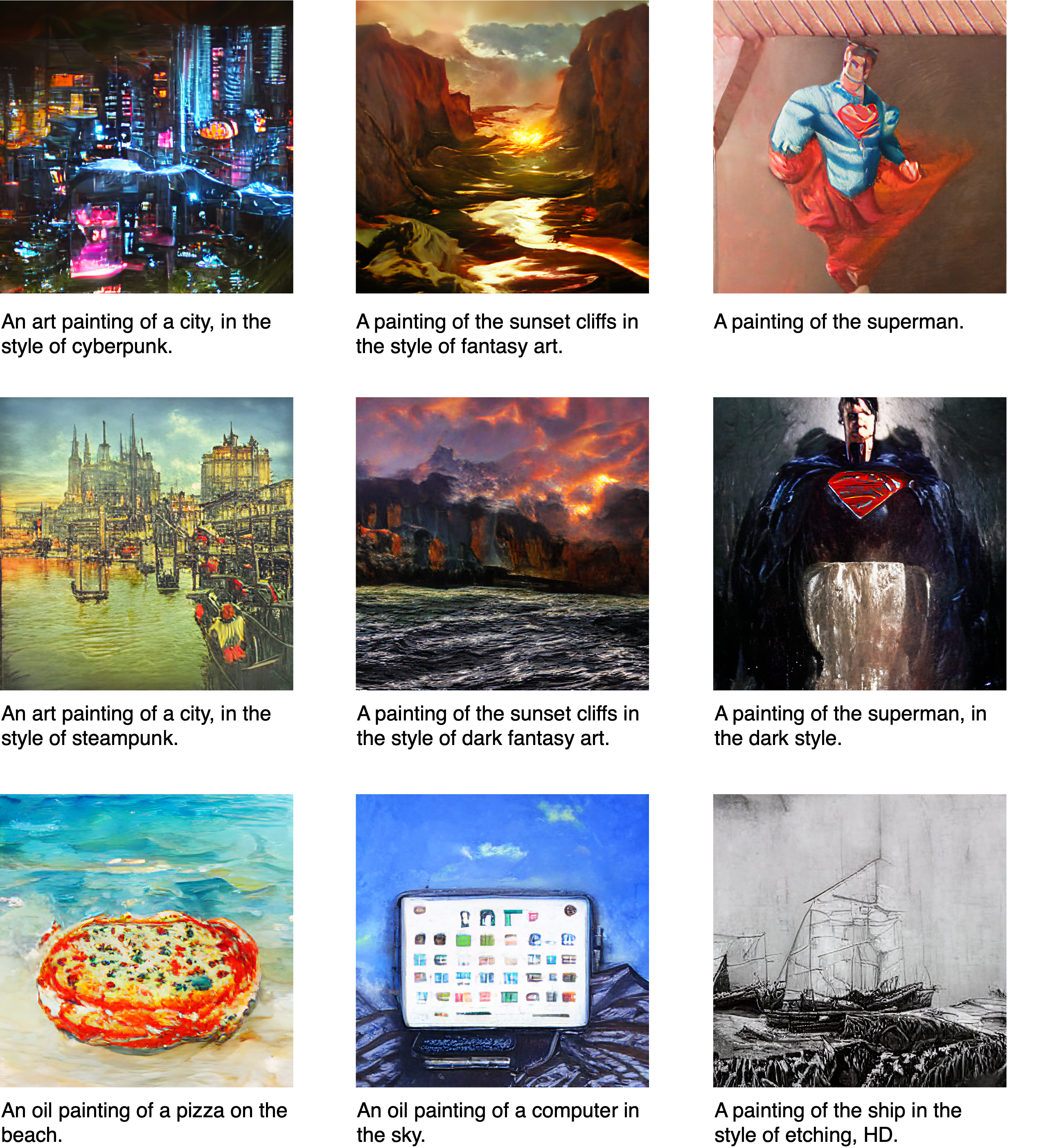}
    \caption{Examples of text-to-image generation. }
    \label{fig:image_gen_samples2}
\end{figure*}

% \begin{figure*}[t]
%     \centering
%     \includegraphics[width=.9\linewidth]{pic/image_gen_cases.pdf}
%     \caption{More samples of text-to-image generation task generated by \ofa~for normal queries.}
%     \label{fig:image_gen_samples1}
% \end{figure*}

% \begin{figure*}[t]
%     \centering
%     \includegraphics[width=.9\linewidth]{pic/counterfactual_images.pdf}
%     \caption{More samples of text-to-image generation task generated by \ofa~for counterfactual queries.}
%     \label{fig:image_gen_samples2}
% \end{figure*}

\begin{figure*}[t]
    \centering
    \includegraphics[width=.9\linewidth]{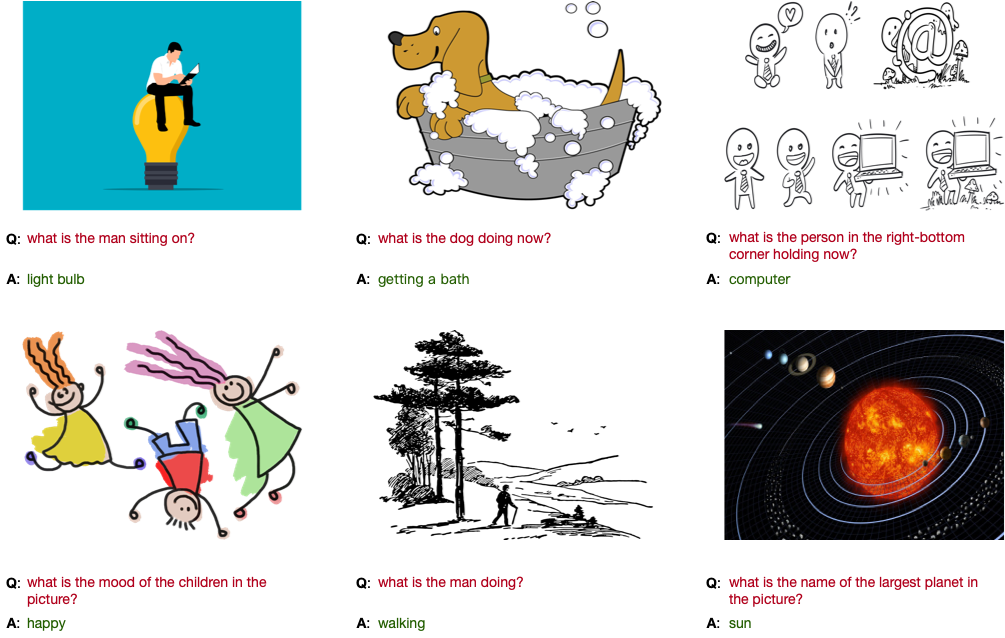}
    \caption{More samples of VQA task on unseen domains. The answers are generated by pretrained \ofa~without finetuning. The datasets used in VQA pretraining task only contain real-world photographs. We present more cases of VQA task on out-of-domain~(non-photographic) images and demonstrate the capability of transferring \ofa~to these unseen domains.}
    \label{fig:ood_case_appendix}
\end{figure*}

\begin{figure*}[t]
    \centering
    \includegraphics[width=.9\linewidth]{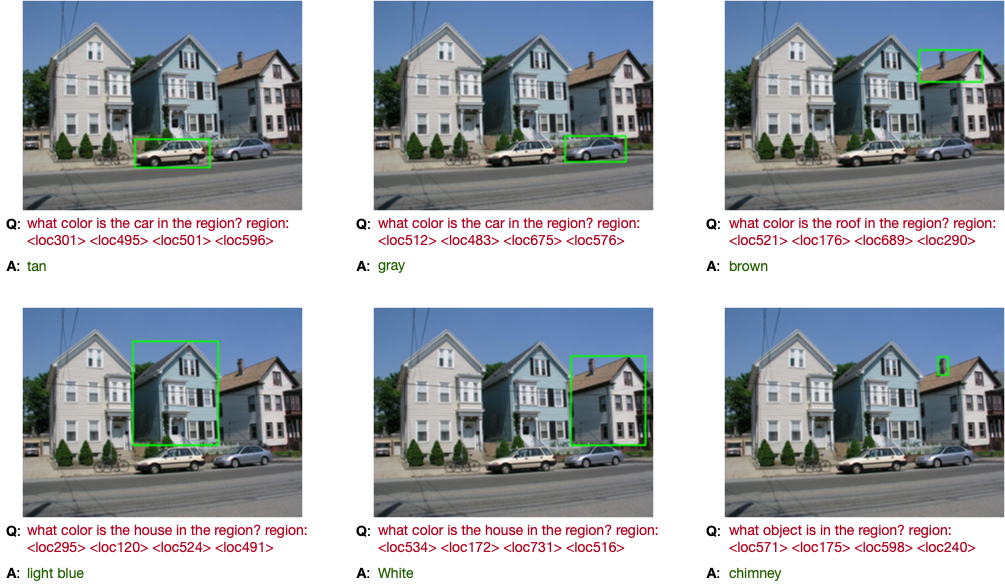}
    \caption{Samples of the unseen grounded question answering task. In this task, the model should answer a question about a particular region in the image. This task is unseen in pretraining. We demonstrate that directly transferring pretrained \ofa~to this new task without finetuning works well.}
    \label{fig:instruction_case_appendix}
\end{figure*}

\begin{figure*}[t]
    \centering
    \includegraphics[width=.9\linewidth]{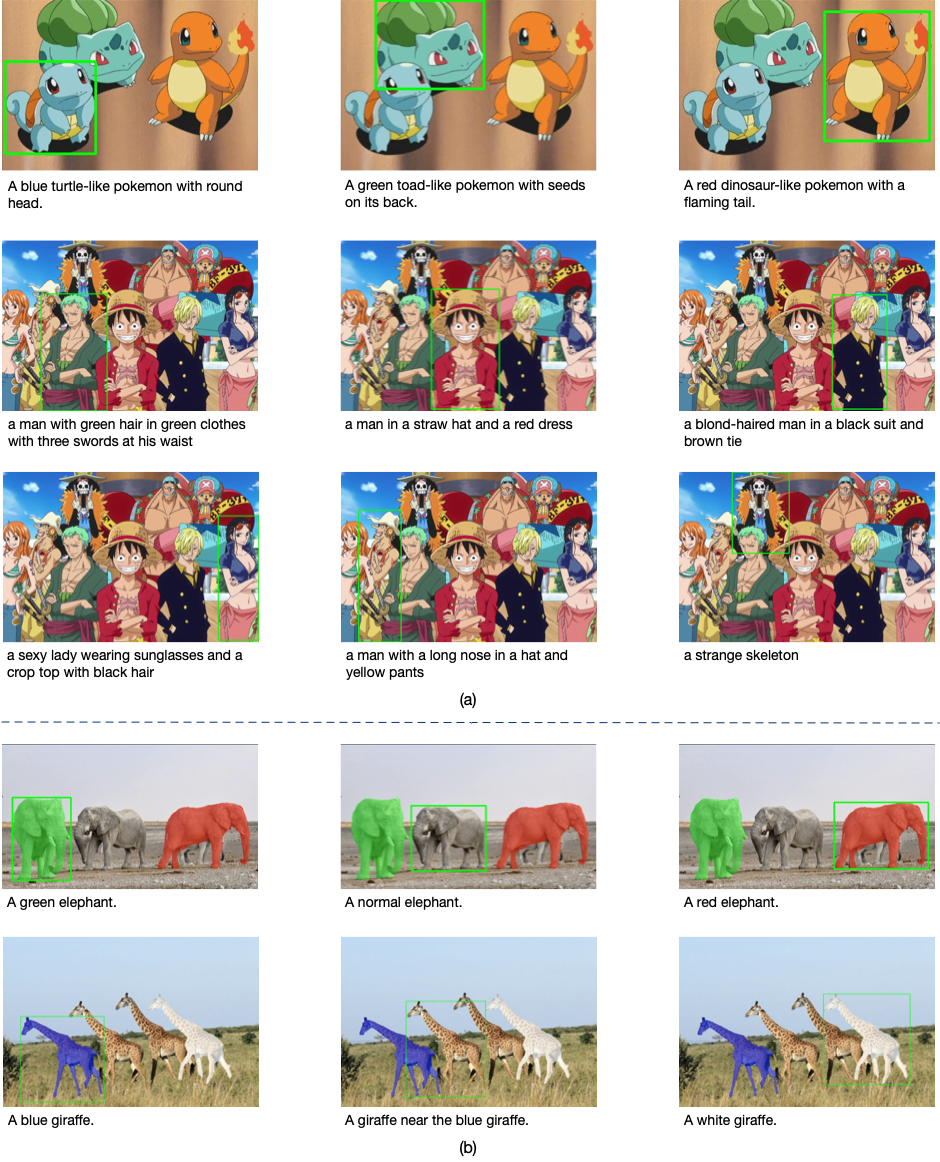}
    \caption{Samples of visual grounding task generated by \ofa~for various unseen domains: (a) anime (the corresponding animations are \textit{Pokemon} and \textit{One Piece}); (b) synthetic images with attribute combinations.}
    \label{fig:vg_samples}
\end{figure*}

\end{document}